\documentclass[journal]{IEEEtran}
\usepackage{graphicx}
\usepackage{amsmath}
\usepackage{algorithmic}
\usepackage{array}
\usepackage[caption=false,font=footnotesize]{subfig}
\usepackage{url}
\usepackage{balance}

\ifCLASSINFOpdf
\else
\fi
\begin{document}
\title{Artifacts Detection and Error Block Analysis from Broadcasted Videos}
\author{Md Mehedi Hasan,~\IEEEmembership{Member,~IEEE,}
        Tasneem Rahman, ~\IEEEmembership{Student Member,~IEEE,}\\
        Kiok Ahn, and Oksam Chae,  ~\IEEEmembership{Member,~IEEE} % stops a space
\thanks{M. M. Hasan is with PBRC, University of South Australia, SA 5095, AUS (e-mail:mdmehedi.hasan@unisa.edu.au).}
\thanks{T. Rahman with SEIT, University of New South Wales, ACT 2610, AUS (e-mail: tasneem.rahman@adfa.edu.au).}
\thanks{K. Ahn and O. Chae are with Department of Computer Engineering, Kyung Hee University, Youngin-si, S.KOR (e-mail: kiok@daum.net; oschae@khu.ac.kr).}
\thanks{}}
%\markboth{IEEE TRANSACTIONS ON CONSUMER ELECTRONICS}
%{Mehedi \emph{\lowercase{et al.}}: Artifacts Detection and Error Block Analysis from Broadcasted Videos}
\maketitle

%%%%%%%%%%%%%%%%%%%%%%%%%%%%%%%%%%%%%%%%%%%%%%%%%%%%%%%%%%%%%%%%%%%%%%%%%%%%%%%%
\begin{abstract}
With  the advancement of IPTV and HDTV technology, previous subtle errors in videos are now becoming more prominent because of the structure oriented and compression based artifacts. In this paper, we focus towards the development of a real-time video quality check system. Light weighted edge gradient magnitude information is incorporated to acquire the statistical information and the distorted frames are then estimated based on the characteristics of their surrounding frames. Then we apply the prominent texture patterns to classify them in different block errors and analyze them not only in video error detection application but also in error concealment, restoration and retrieval. Finally, evaluating the performance through experiments on prominent datasets and broadcasted videos show that the proposed algorithm is very much efficient to detect errors for video broadcast and surveillance applications in terms of computation time and analysis of distorted frames.
\end{abstract}
%%%%%%%%%%%%%%%%%%%%%%%%%%%%%%%%%%%%%%%%%%%%%%%%%%%%%%%%%%%%%%%%%%%%%%%%%%%%%%%%

% Note that keywords are not normally used for peer review papers.
\IEEEpeerreviewmaketitle

%%%%%%%%%%%%%%%%%%%%%%%%%%%%%%%%%%%%%%%%%%%%%%%%%%%%%%%%%%%%%%%%%%%%%%%%%%%%%%%%
\section{Introduction}
\label{intro}

\par
\begin{figure}
\centering
  \includegraphics[width = 3.4 in]{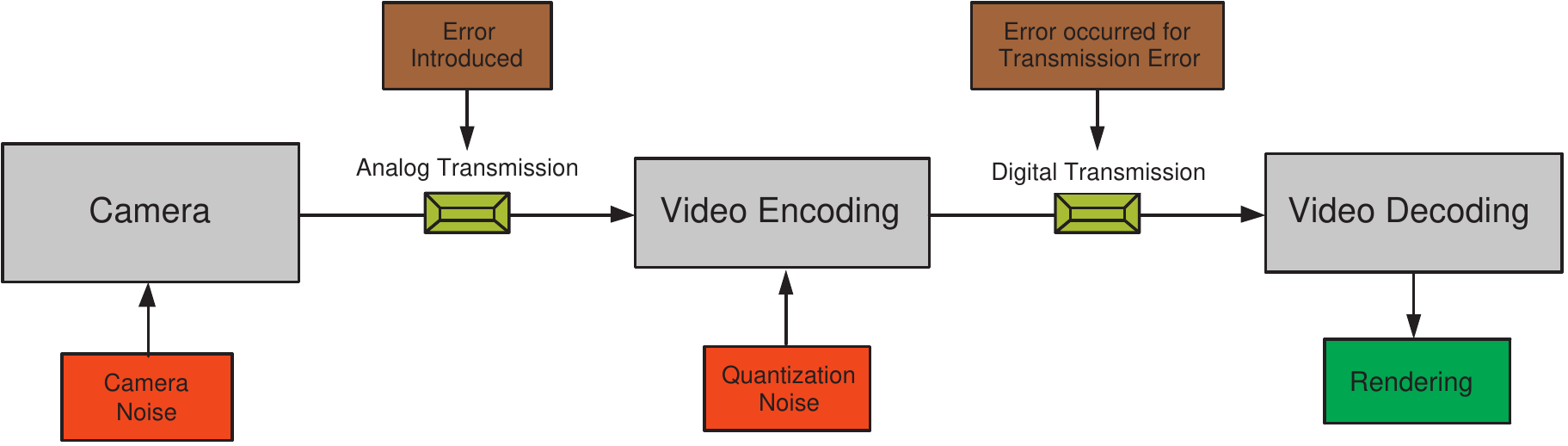}
\caption{Noise and error model for video broadcasting and surveillance systems}
\label{Fig1_Surveillance}       % Give a unique label
\end{figure}

 With the rapid development of the application of video surveillance and broadcast systems, the evaluation of video quality becomes an emerging research. Error detection is an important criterion to measure the quality of images/videos transmitted over unreliable networks particularly in the wireless channel. At the time of acquisition and transmission, video frames are always distorted by various artifacts. In a real system, noises are mainly introduced by the camera and the quantization step of decoding process as showed in Fig.\ref{Fig1_Surveillance}. But the distortions \cite{Wang_errorcontrol98} occur when the videos are transmitted through analog or digital medium. Some errors may be introduced when the analog video signal transmits in wired channel, but in wireless communication it cannot be ignored as occurring frequently. So, structure-oriented video distortions detection significantly impacts the effectiveness of video processing algorithms.

\par
\begin{figure}
\centering
  \includegraphics[width = 3.4 in]{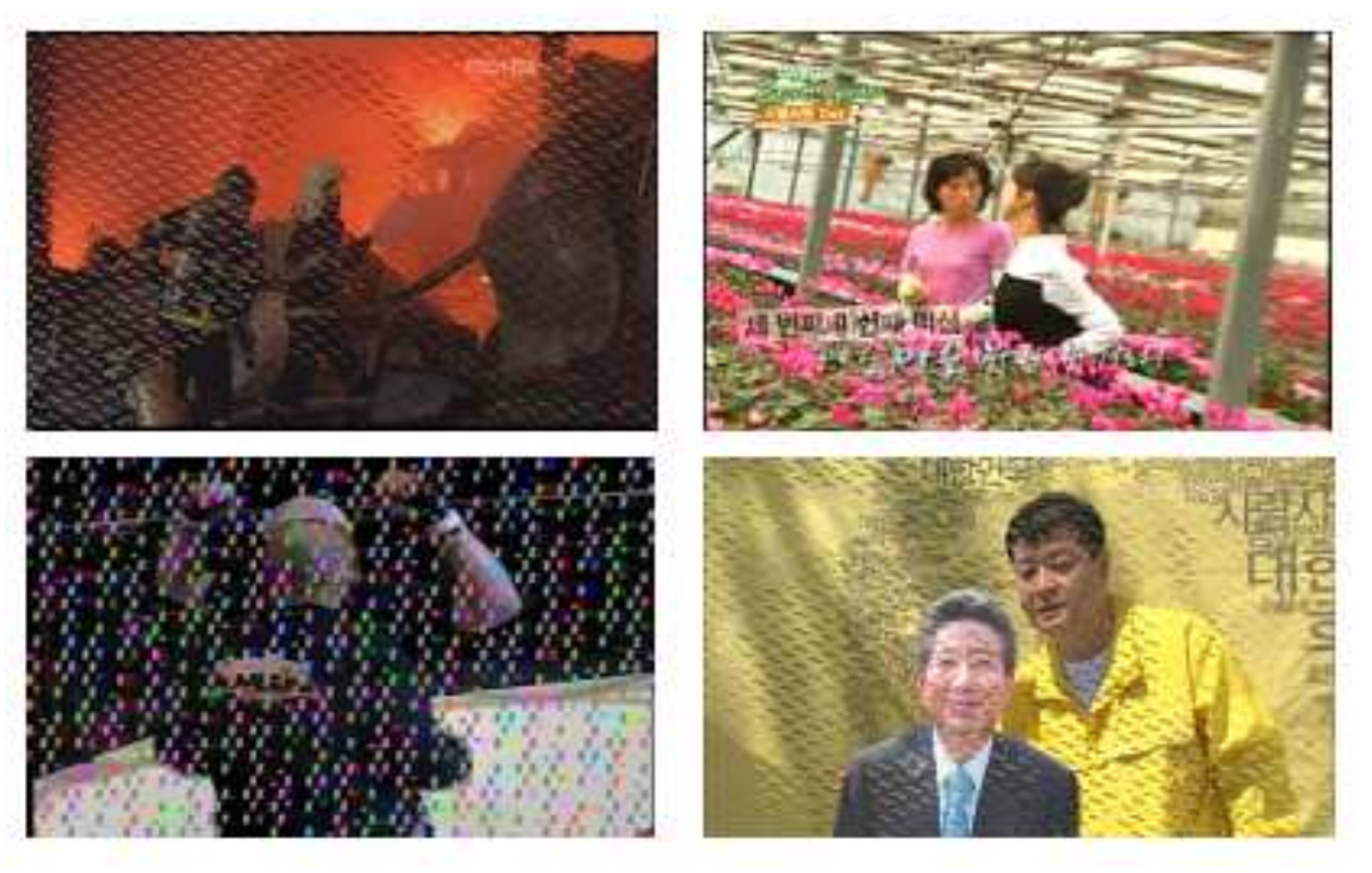}
\caption{Different kind of video artifacts occurred during transmission (Courtesy: KBS)}
\label{Fig2_Broadcast1}       % Give a unique label
\end{figure}

 There are many potential methods of measuring discrete cosine transform (DCT) based codec degradations involve directly examining the coarseness of the compressed video stream at the time of quantization scaling. The B-DCT scheme \cite {Liu_EDCT02} takes into account the local spatial correlation property of the images by dividing the image into 8 by 8 block of pixels. In DCT each block pixels are treated as single entity and coded separately. A slight change of luminance in border area can cause a step in the decoded image if the neighboring blocks fall into different quantization intervals. Therefore, the decompressed image and video exhibits various kind of artifacts. One of the most obtrusive artifacts is the "Blocking Artifact"\cite{Stumuller_lossy00}. But in broadcasting system, distortions are not always taking place in block by block basis. Sometimes the contents (audio and video) of broadcast system include distortions by bad cassette header, defects in encoding processes and devices, poor tape aging and storage, Non-Linear Editing error \cite{Stumuller_lossy00} and so on \cite{Wang_errorcontrol98}. Fig.\ref{Fig2_Broadcast1} shows different kind of video errors occurred during transmission. It is also observed that these errors are not sustaining in a block based manner and cannot be defined like compression based artifacts. Now-a-days, it is a challenging research issue to measure both distortions which occur at the time of compression and broascasting in the same manner.

%%%%%%%%%%%%%%%%%%%%%%%%%%%%%%%%%%%%%%%%%%%%%%%%%%%%%%%%%%%%%%%%%%
%%%%%%%%%%%%%%%%%%%%%%%%%%%%%%%%%%%%%%%%%%%%%%%%%%%%%%%%%%%%%%%%%%

\par In this paper, we focus towards the development of a real-time video quality check system.To overcome the above issues we first propose a measurement system for compression and broadcasting related artifacts and then analyze the distortion patterns occurred during its transmission over wireless channels. Firstly, The proposed system is achieved by a measurement of various artifacts of videos by analyzing the distribution of local properties of image signals like dominant edge magnitude and direction. Then we propose a method to analyze the distortion pattern or block errors in video frames that occurred during its transmission.

%%%%%%%%%%%%%%%%%%%%%%%%%%%%%%%%%%%%%%%%%%%%%%%%%%%%%%%%%%%%%%%%%%
%%%%%%%%%%%%%%%%%%%%%%%%%%%%%%%%%%%%%%%%%%%%%%%%%%%%%%%%%%%%%%%%%%

%%%%%%%%%%%%%%%%%%%%%%%%%%%%%%%%%%%%%%%%%%%%%%%%%%%%%%%%%%%%%%%%%%%%%%%%%%%%%%%%%%%%%%%%%%%%%%%%%%
%%%%%%%%%%%%%%%%%%%%%%%%%%%%%%%%%%%%%%%%%%%%%%%%%%%%%%%%%%%%%%%%%%%%%%%%%%%%%%%%%%%%%%%%%%%%%%%%%%

\section{Previous Research Works}
\label{prev}

\par The issue of detecting video artifacts is closely related with the field of video quality measurement which has been widely studied. Many video quality metrics in the research field use a standard defined by the International Telecommunications Union (ITU) technical paper "Methodology for the subjective assessment of the quality of television pictures" \cite {Quality_tv03}. This work conducted a series of subjective tests which tabulated the mean opinion scores (MOS) against a database of videos.Out of several video metrics created \cite {Wolf_quality02} \cite{Feng00} \cite{Watson_dvq01} \cite {Loke_multimedia06} , one of the better performing metric was the National Telecommunications and Information Administration (NTIA) video quality metric (VQM) \cite {Wolf_quality02}, which scores relatively better over a wide range of videos. The VQM metric used a set of weighted-parameters on several components such as image blurriness, color, and presence of blockiness. In another work \cite {Loke_multimedia06}, the results showed that video quality metrics in general did not perform well when restricted to the videos with low bit ranges. Although there is research on the effect of the video artifacts toward the overall video quality, there has been limited research on the individual artifacts itself.

\par A previous work by Qi et al. was done as a subjective test which measured the effect of frame freezing and frame skipping on the video quality \cite {Qi_freeze06}. In this work, the freeze artifacts and loss artifacts are inserted randomly into parts of the sequences. However, the results of the experiment still aimed at determining the overall video quality, instead of the individual artifacts. In another work by Lu et al., the effect of frame dropping and blurriness on the overall video quality is measured, to examine the relative strength of each artifact to each other \cite{Lu_drop07}. The various factors that contributed to the perceived blur effect included the motion, image contrast and orientation of the distorted videos. While many metrics and studies aim at investigating the effects of blockiness artifacts on the overall quality of the video sequence, there are relatively few tests trying to quantify the presence of the blockiness artifact itself \cite {Wu_impair97}\cite{Malvar_LOT89}\cite{Paek_POCS98}\cite{Wang_novel98}. Most of these works are related to the video processing field, which try to reduce the effects of blockiness present, and cannot be used to detect the blockiness that is induced through hardware defects. In image coding techniques, Chou et. al. \cite{Chou_JND95} addressed a key concept of perceptual coding considering human visual system, namely just-noticeable-distortion (JND). JND is a function of local signal properties, such as background intensity, activity of luminance changes, dominant spatial frequency and changes in edge gradients. Once the JND profile of an image is obtained, the energy of the perceptible distortion like blockiness can be measured. But this kind of HVS \cite{Karunasekera_HVS95}\cite {Liu_noref08}\cite {Zhen_vision02} measurement system is computationally expensive and cannot be applicable for fast real time cases.

\par For extracting the features from images or videos pixel based and edge based methods are used \cite{Zhang_conceal04}. In pixel based method to consider HVS various kinds of masking is used like, texture masking, luminance masking, just noticeable distortion profile etc. In these pixel-based methods, pixel intensity shows high sensitivity to illumination variation and noise, and thereby degrades the overall performance of the detection result \cite{Hsu_change84}. With compare to pixel intensity, edge feature is more robust which achieves two major advantages; (a) Less sensitive to illumination variation and noise and (b) It requires less computation than analysis of entire grayscale image in pixel intensity based methods.Though these methods show more robustness with compare to region based methods, they still face challenges with poor representation and improper utilization of edges in case of moving object detection. As a result, existing edge based methods experience poor detection result in dynamic environment which eventually affects the further higher level processing of video surveillance and broadcasting systems such as moving object segmentation, tracking, shape recognition, error detection and noise estimation.

\par To overcome all the limitations, we propose a metric to detect damaged frame by considering the contextual information, such as their consistency and edge continuity. To achieve HVS we incorporate light weighted edge gradient magnitude information for video artifacts. According to the statistical information the distorted frames are then estimated based on the characteristics of their surrounding frames. Then we generate a criteria function to detect the distorted frame from sequence of video frames. Secondly, we propose a method to analyze the distortion pattern or block errors in video frames. To achieve real-time performance we quantize edge gradient phase information of the image and histogram of the quantized values. By making the artifact characteristics, we can filter out background pixels and compression based patterns. Then use the prominent artifacted texture patterns to classify them in different block errors and analyze them for error concealment. So, through accumulating histogram-based edge gradient information we can achieve height, width, shape and rotation of the distortion patterns of the video frames and analyze the distorted content pattern not only in video error detection application but also in error concealment, restoration and retrieval.

%%%%%%%%%%%%%%%%%%%%%%%%%%%%%%%%%%%%%%%%%%%%%%%%%%%%%%%%%%%%%%%%%%%%%%%%%%%%%%%%%%%%%%%%%%%%%
%%%%%%%%%%%%%%%%%%%%%%%%%%%%%%%%%%%%%%%%%%%%%%%%%%%%%%%%%%%%%%%%%%%%%%%%%%%%%%%%%%%%%%%%%%%%%

\section{Distortion Measure and Error Frame Detection}
\label{prop1}

\par
\begin{figure*}
\begin{center}
  \includegraphics[width = 5.4 in]{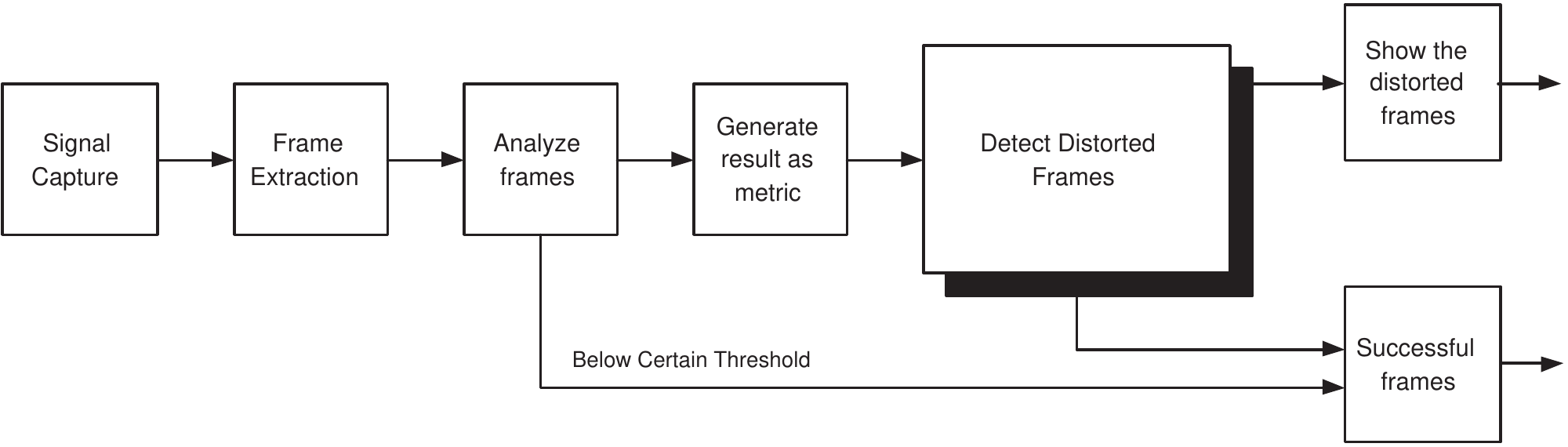}
\end{center}
\caption{Flow diagram of the proposed System}
\label{Fig_Chapter3_Proposed}
\end{figure*}

 In our approach we assume that our video is reference free and so we have to blindly measure the artifacts of videos or images. The overall proposed system is shown in Fig.\ref{Fig_Chapter3_Proposed}. Firstly, the compressed video input is given to a signal module to measure the degraded picture quality and analyze the measure to generate a distortion metric. Then the content of each frame is analyzed to measure the artifacts of that frame.

%%%%%%%%%%%%%%%%%%%%%%%%% Chap3_Fig.16 starts here %%%%%%%%%%%%%%%%%%%%%%%%
%%%%%%%%%%%%%%%%%%%%%%%%% Chap3_Fig.16 ends here %%%%%%%%%%%%%%%%%%%%%%%%%%

%
\subsection{Gradient Magnitude Accumulation}
\label{Sec_Chapter3_MeasureDetect}
\begin{figure}
\centering
\includegraphics[width = 3.5 in]{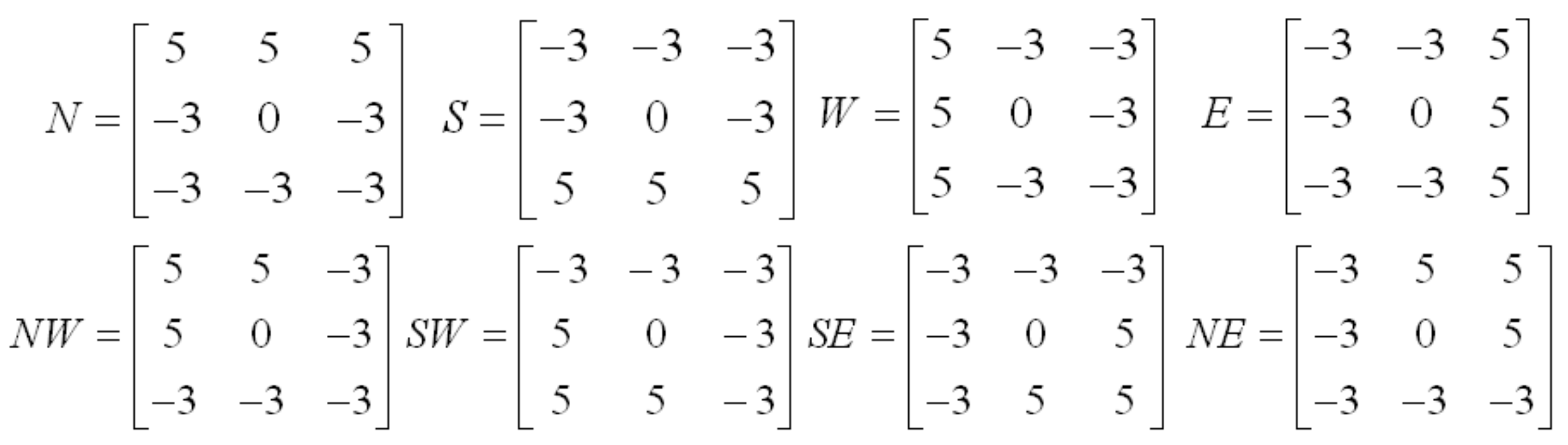}
\label{Fig_Chapter2_Kirsch}
\includegraphics[width = 3.4 in]{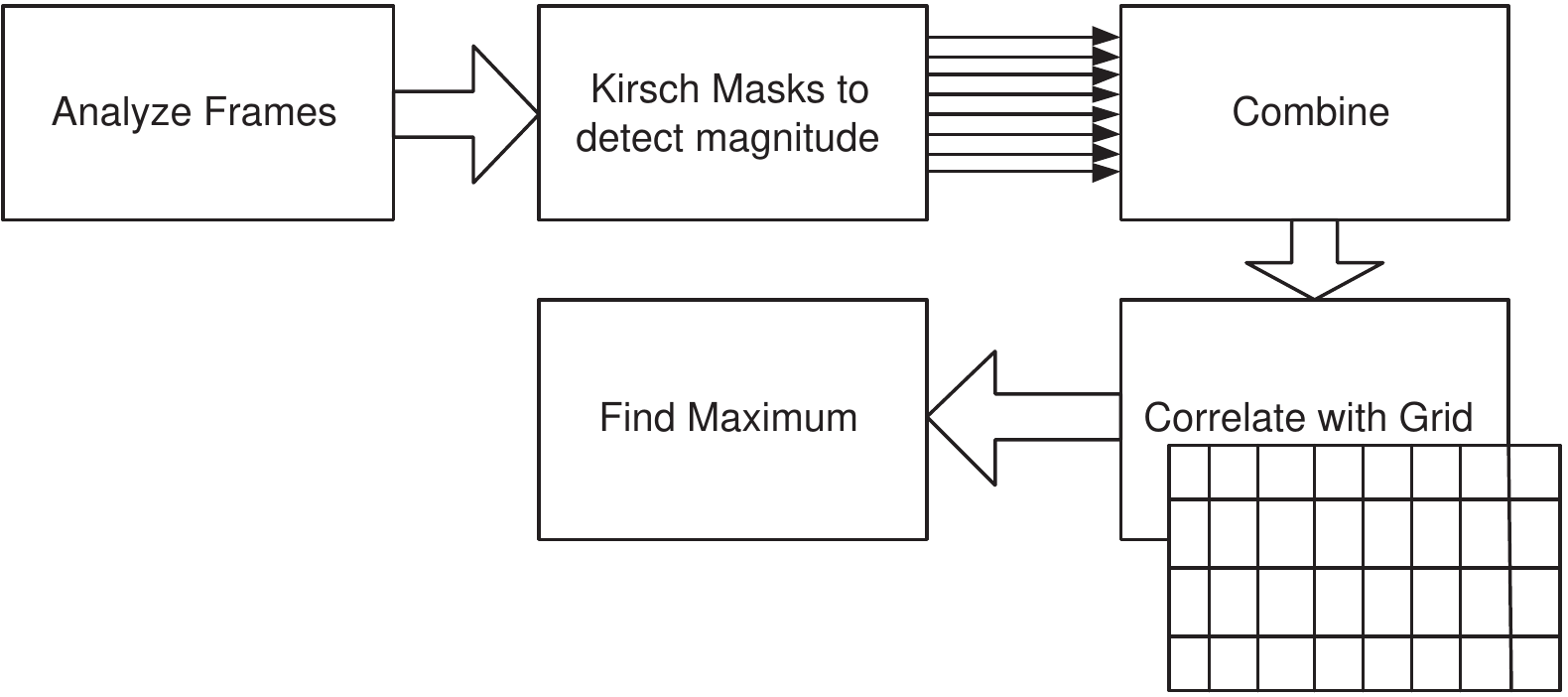}
\label{Fig_Chapter3_Proposed1}
\caption{Eight directional Kirsch Mask and Correlation with grid}
\label{Fig_Chapter3_Prop}
\end{figure}

A frame of a video signal representing an image is captured and converted, if necessary into luminance and color components. Traditionally, one or more of the components is analyzed by appropriate vertical and horizontal edge enhancement filtering. But if we consider the high gradient information in eight directions and use this information for the distortion measurement, we will gain apparently more accurate results. So, to consider the different directional information we use Kirsch Masks \cite{Pratt07} in eight directions as showed in Fig.\ref{Fig_Chapter3_Prop}. The edge magnitude is equal to the maximum value found by the convolution of each mask with the image. The edge direction is defined by the mask that produces the maximum magnitude.

\par The $ g(x,y) $ across the pixel at $(x,y) $  is determined by calculating the edge changes in eight directions and eight operators, $ G_{k}(x,y)$ , for $ k = 1, 2,¡¦,¡¦,8 $ and $ i,j = 1,2,3, $ are employed to perform the calculation.

\begin{equation}
g(x,y)=\max_{k=1,...,8}^{}{}\left | grad_{k}(x,y) \right |
\label{Equ_Chapter3_Proposed1}
\end{equation}

\begin{equation}
grad_{k}(x,y)=\sum_{i=1}^{3}\sum_{j=1}^{3}P(x-2+i,y-2+i).G_{k}(i,j)
\label{Equ_Chapter3_Proposed2}
\end{equation}

\par for $0\leq x\leq H $ and $0\leq y\leq W $.
\newline
\par Where $ p(x,y) $  denotes the pixel at $(x,y) $. $H $ and $W $  are  the height and width of the frame consecutively. The resulting edges are correlated with an infinite grid having boundaries corresponding to the block boundaries used in the video compression. Fig.\ref{Fig_Chapter3_Prop} shows the consequence after using Kirsch masks to detect the edges. Optionally, a second correlation may be made with boundaries slightly different than the block boundaries used in the video compression, with this result being subtracted from the first value. Further the locations of the compression block boundaries may be detected by observing where the maximum correlation value occurs. The resulting correlation results are proposed to generate a picture quality rating for the image which represents the amount of human-perceivable block degradation that has been introduced into the proposed video signal.

%%%%%%%%%%%%%%%%%%%%%%%%% Chap4_Fig.23 starts here %%%%%%%%%%%%%%%%%%%%%%%%
%%%%%%%%%%%%%%%%%%%%%%%%% Chap4_Fig.23 ends here %%%%%%%%%%%%%%%%%%%%%%%%%%

%%%%%%%%%%%%%%%%%%%%%%%%% Chap3_Fig.19 starts here %%%%%%%%%%%%%%%%%%%%%%%%
\begin{figure*}
\begin{center}
  \includegraphics[width = 5.5 in]{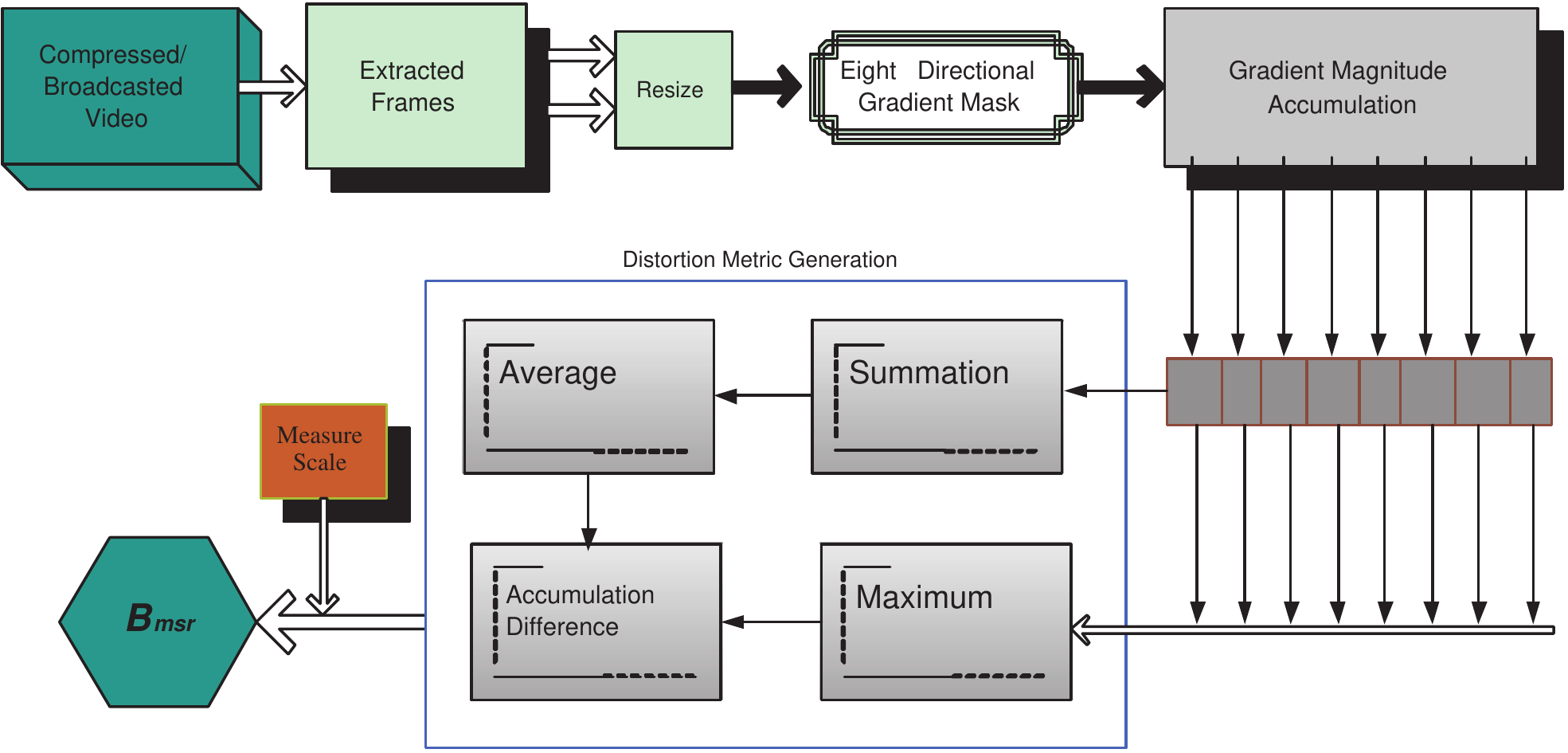}
\end{center}
\caption{Picture quality measurement system}
\label{Fig_Chapter3_ProposedFinal}
\end{figure*}
%%%%%%%%%%%%%%%%%%%%%%%%% Chap3_Fig.19 ends here %%%%%%%%%%%%%%%%%%%%%%%%%%

%%%%%%%%%%%%%%%%%%%%%%%%% Chap4_Fig.20 starts here %%%%%%%%%%%%%%%%%%%%%%%%
\begin{figure*}
\begin{center}
  \includegraphics[width = 5.5 in]{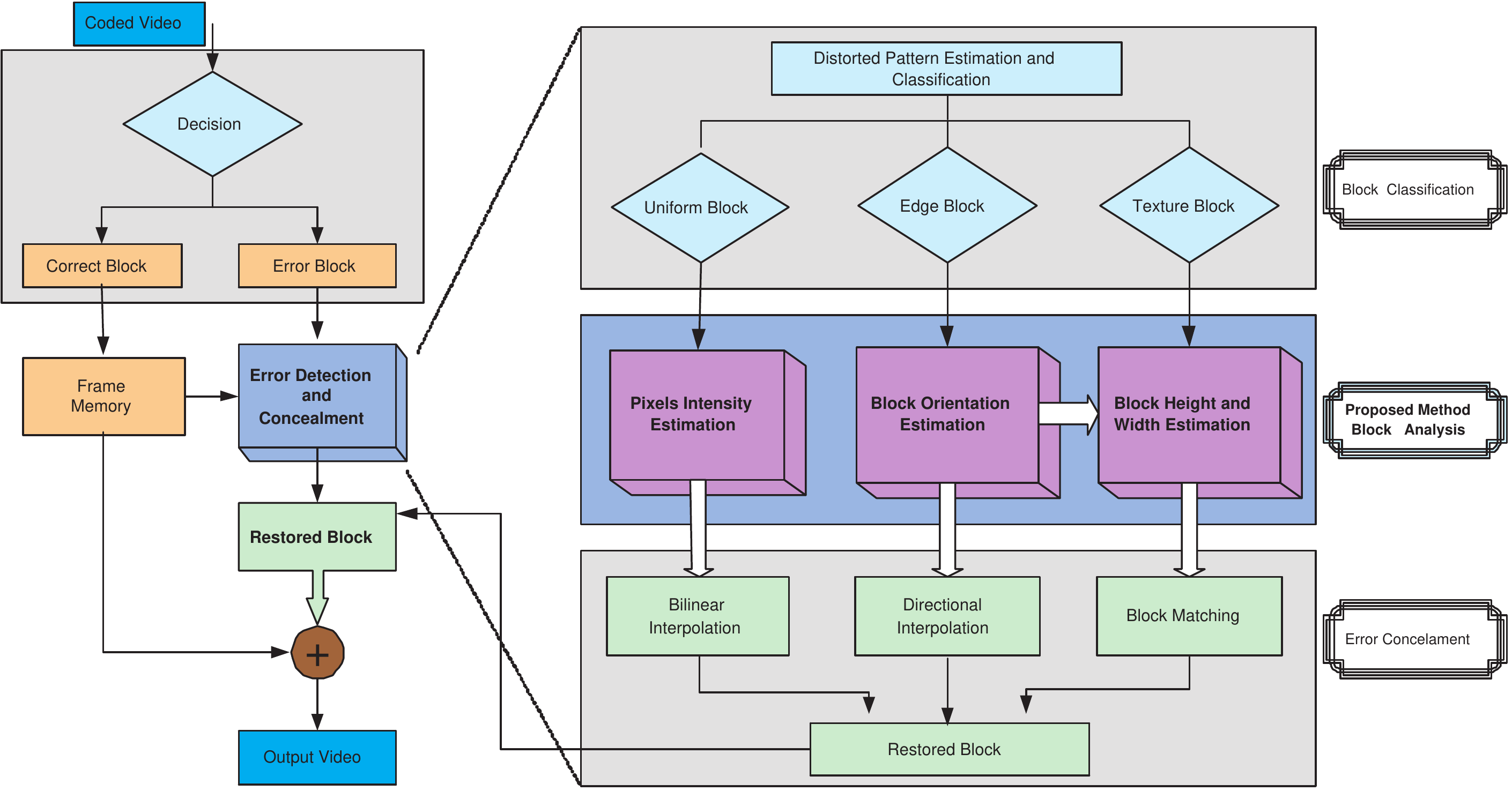}
\end{center}
\caption{Framework of the Overall system for Error Classification, Analysis and Concealment}
\label{Fig_Chapter4_Framework}
\end{figure*}
%%%%%%%%%%%%%%%%%%%%%%%%% Chap4_Fig.20 ends here %%%%%%%%%%%%%%%%%%%%%%%%%%

\subsection{Distortion Metric Generation}
\par To generate the distortion metric for real time systems we have to consider a faster and efficient approach to measure and also detect the location of the distortion occurred in a frame. Many of the existing algorithms are based on HVS. But these kinds of algorithms are not faster enough to use in real time. So, we have proposed a simpler approach. Fig.\ref{Fig_Chapter3_ProposedFinal} shows a block diagram of a picture quality measurement system. The steps to detect block error after using the kirsch masks is shown in the algorith as below:

%%%%%%%%%%%%%%%%%%%%%%%%% Algorithm 1 starts here %%%%%%%%%%%%%%%%%%%%%%%%%%

\begin{math}
\\
 1.\alpha \leftarrow absolute[grad_{k}(x,y)]\\
 2.\Omega \leftarrow \alpha\; (20:1900, 20:1060) \; //Optional\; Clipping \\
 3.\Delta \leftarrow Block \; Size \\
 4.\Theta \leftarrow array\left (  0,\Delta \right )= 0 \\
 5.\beta _{C8}\leftarrow floor\, (W/\Delta )\ast \Delta -\Delta\\
 // Round \: width \; down \; to\:   nearest\times \Delta \\
 6.for\; i\leftarrow 1\; to\; length\; [H] \\
 7.\; \;do \; \; for\; j\leftarrow 0\; to\; \Delta \\
 8.\; \; \; do \; \; for\; k \leftarrow 0\; to\; \Delta \\
 9. \; \; \; \; then\; \Theta(k)\: = \: \Theta(k)+V(i,j+k)\\
 10. \; \; \; \;do \; \; if \; j < \beta_{C8}  \; \; go to \;\;  step \; 2\\
 \end{math}

%%%%%%%%%%%%%%%%%%%%%%%%% Algorithm 1 ends here %%%%%%%%%%%%%%%%%%%%%%%%%%

\par Here we assume the block size as $ \Delta \times \Delta $. We add the values into $\Delta $ buckets $ \Theta (k) $ with each bucket containing the total of all the columns by modulo eight (bucket: $1$ contains the sum of columns $ 1, 9, 17, 25$ bucket: $2$ contains the sum of columns $ 2, 10, 18, 26) $. And finally we achieve the measurement values by the following Equation \ref{Equ_Chapter3_Proposed3} and \ref{Equ_Chapter3_Proposed4}:

\begin{equation}
\Theta_{\Delta } (k)=\left [ \sum_{k =0}^{\Delta }\Theta (k)) \right ]/\Delta
\label{Equ_Chapter3_Proposed3}
\end{equation}

\begin{equation}
B_{msr}(n)=[MAX(\Theta {k})-\Theta _{\Delta }(k)]\times \prod
\label{Equ_Chapter3_Proposed4}
\end{equation}

\par The average values which are acquired from Equation \ref{Equ_Chapter3_Proposed3} are subtracted from the maximum values of the buckets and multiplied by $\prod  $, a scale factor which is a constant value in \ref{Equ_Chapter3_Proposed4} and which range can be achieved as the Picture Quality Rating (PQR) produced by a device like PQA200. Nevertheless, the maximum $ \Theta (k) $ value is also indicates where the compression block boundary is. It is also output to the report module. To improve the algorithm accuracy, measurement can be changed correlating with the block or macro block spacing and then simultaneous calculation using a non-block correlated kernel size. This significantly reduces false positives in blockiness that a noisy image might otherwise produce. The location of the compressed blocks whether they are human-perceivable or not, may be determined and separately reported.

\subsection{Detection of Distorted Frame}
To compute the distortion measure of every frame we compare deviation with the previous frame. If the value is within a certain threshold value then it is considered as successful undistorted frame. Otherwise it is consider as distorted frame and forwarded to next report results module. First we consider previous frames matrices and then compute the mean of the frames and the standard deviation of the frames.

\begin{equation}
F_{r}=\sum_{n=0}^{N}B_{msr}(n)
\label{Equ_Chapter3_Proposed5}
\end{equation}

\begin{equation}
M_{i}=\frac{F_{r}}{n}=\frac{\sum_{n=0}^{N}B_{msr}(n)}{n}
\label{Equ_Chapter3_Proposed6}
\end{equation}

\begin{equation}
\sigma _{i}=\sqrt{\sum_{n=0}^{N}[B_{msr}(n)-M_{i}]^{2}}
\label{Equ_Chapter3_Proposed7}
\end{equation}

\par Here $N$ is the number of frames we want to consider and also $M_{i} $ and $\sigma _{i} $  are the mean and standard deviation respectively. After computing the mean and standard deviation we have to consider how much deviation we allow the frame to be considered as distorted frame. In our proposal, after extensive experiment on various video frames we observe that the desirable condition of a frame to be considered as distorted if:

\begin{equation}
\left [ \frac{(\sigma _{i}-\sigma _{i-1})}{(B_{msr}(n)-B_{msr}(n-1))} \right ]\geq \beta^{2}
\label{Equ_Chapter3_Proposed8}
\end{equation}

\par Here $\beta$ is the error detection criteria function, which varies for different kind of dynamic environment.

\section{Spatial Error Block Analysis}
\label{prop2}

\par
\begin{figure}
\begin{center}
  \includegraphics[width = 3.5 in]{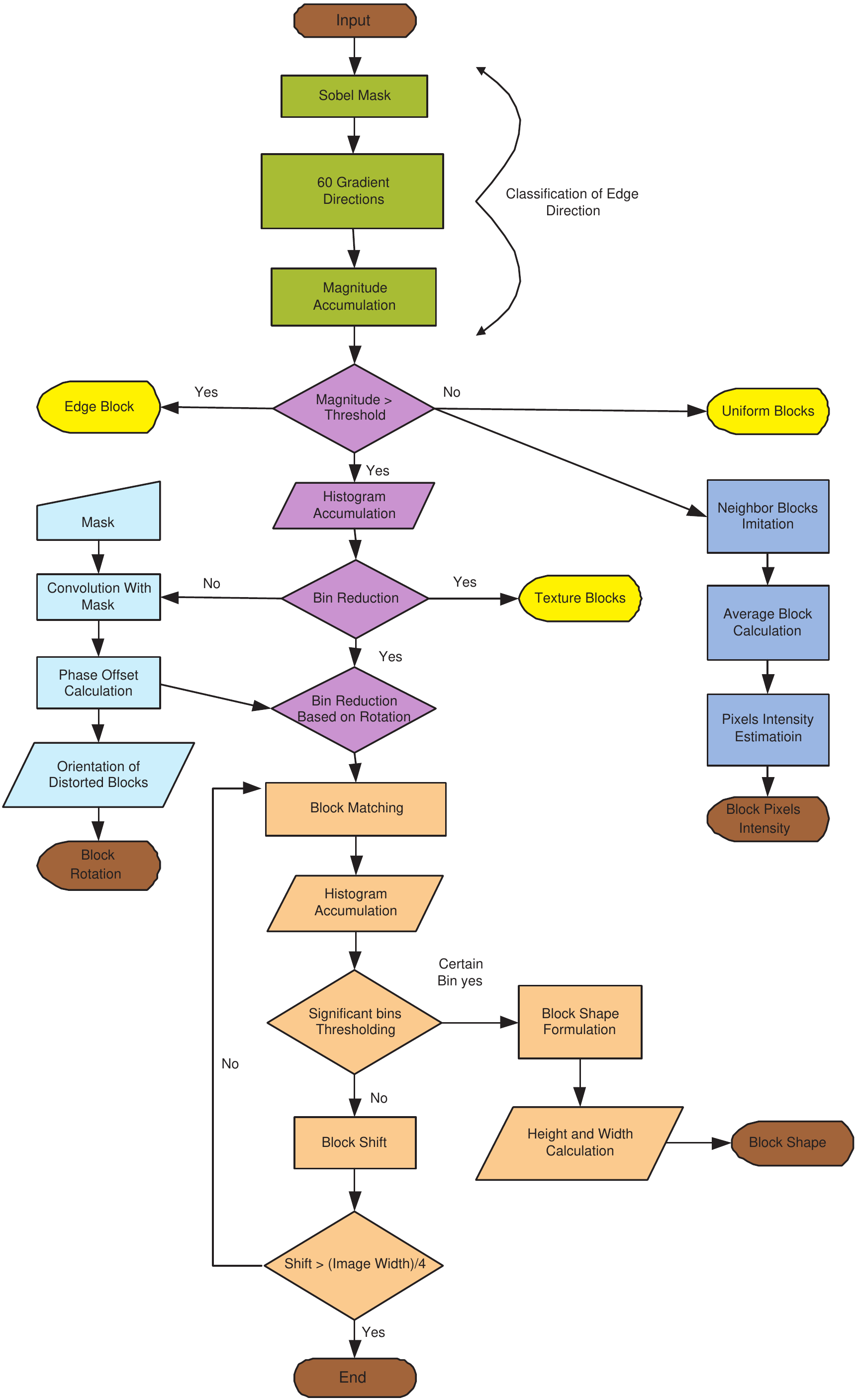}
\end{center}
\caption{Flowchart of our proposed Spatial Error Block Analysis(SEBA) }
\label{Fig_Chapter4_Flowchart}
\end{figure}

 The error block (EB) classification has been proposed by Zhang et. al proposed in \cite{Zhang_conceal04}. But we extend his system by our block analysis method, which is very much important for gaining the accuracy of error concealment and is showed in Fig.\ref{Fig_Chapter4_Framework}. And then the EB is reconstructed by appropriate methods to the appropriate category. In our approach the proposed EC algorithm consists of, i.e., the block classification module , block analyzing and concealment module. The flowchart of the analyzing process is shown in the Fig.\ref{Fig_Chapter4_Flowchart}.

%%%%%%%%%%%%%%%%%%%%%%%%% Chap4_Fig.21 starts here %%%%%%%%%%%%%%%%%%%%%%%%
%%%%%%%%%%%%%%%%%%%%%%%%% Chap4_Fig.21 ends here %%%%%%%%%%%%%%%%%%%%%%%%%%

\subsection{Classification of Edge direction}
\label{Sec_Chapter4_EdgeClassification}

\begin{figure}
\centering
\includegraphics[width = 1.5in]{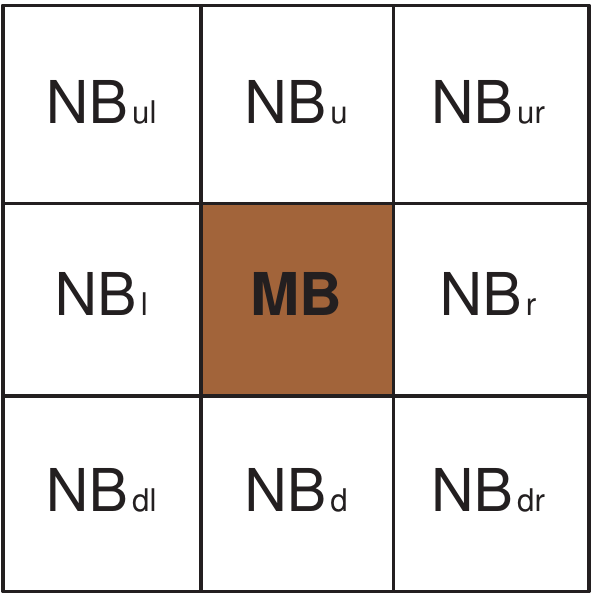}
\label{Fig_Chapter4_Block}
\hspace{0.05in}
\includegraphics[width = 1.5in]{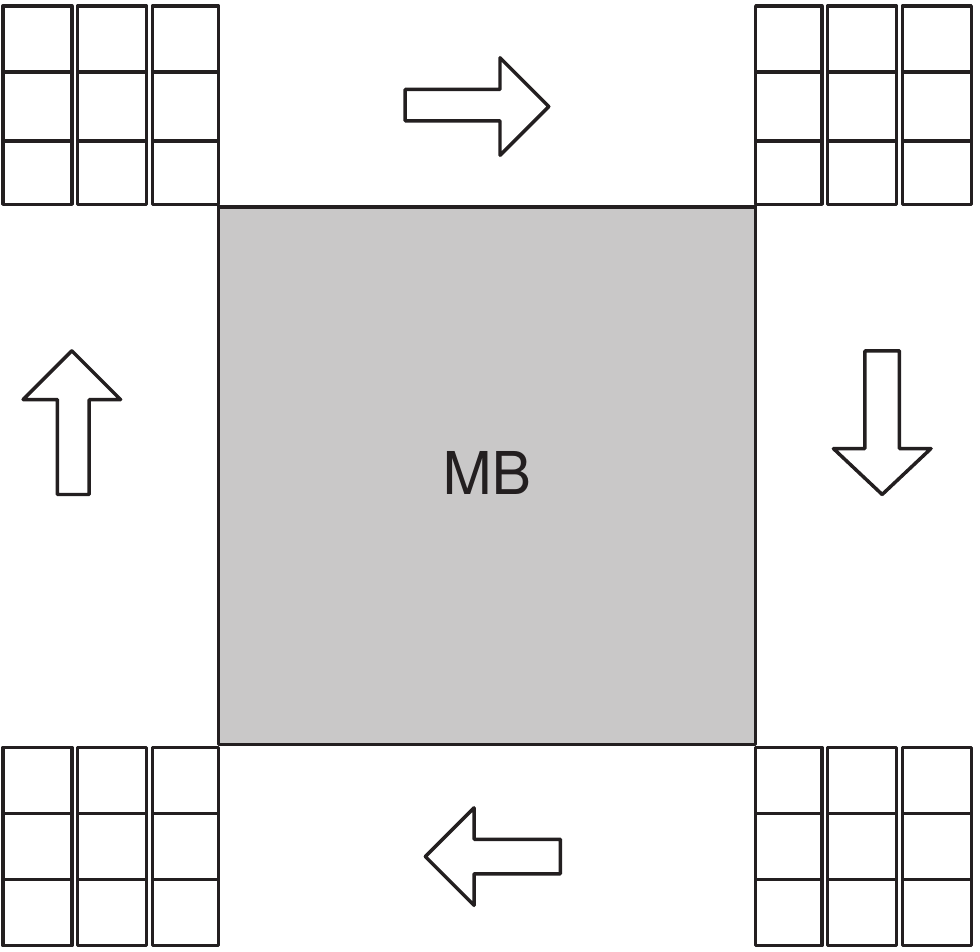}
\label{Fig_Chapter4_Isoblock}
\hspace{0.05in}
\includegraphics[width = 2.5in]{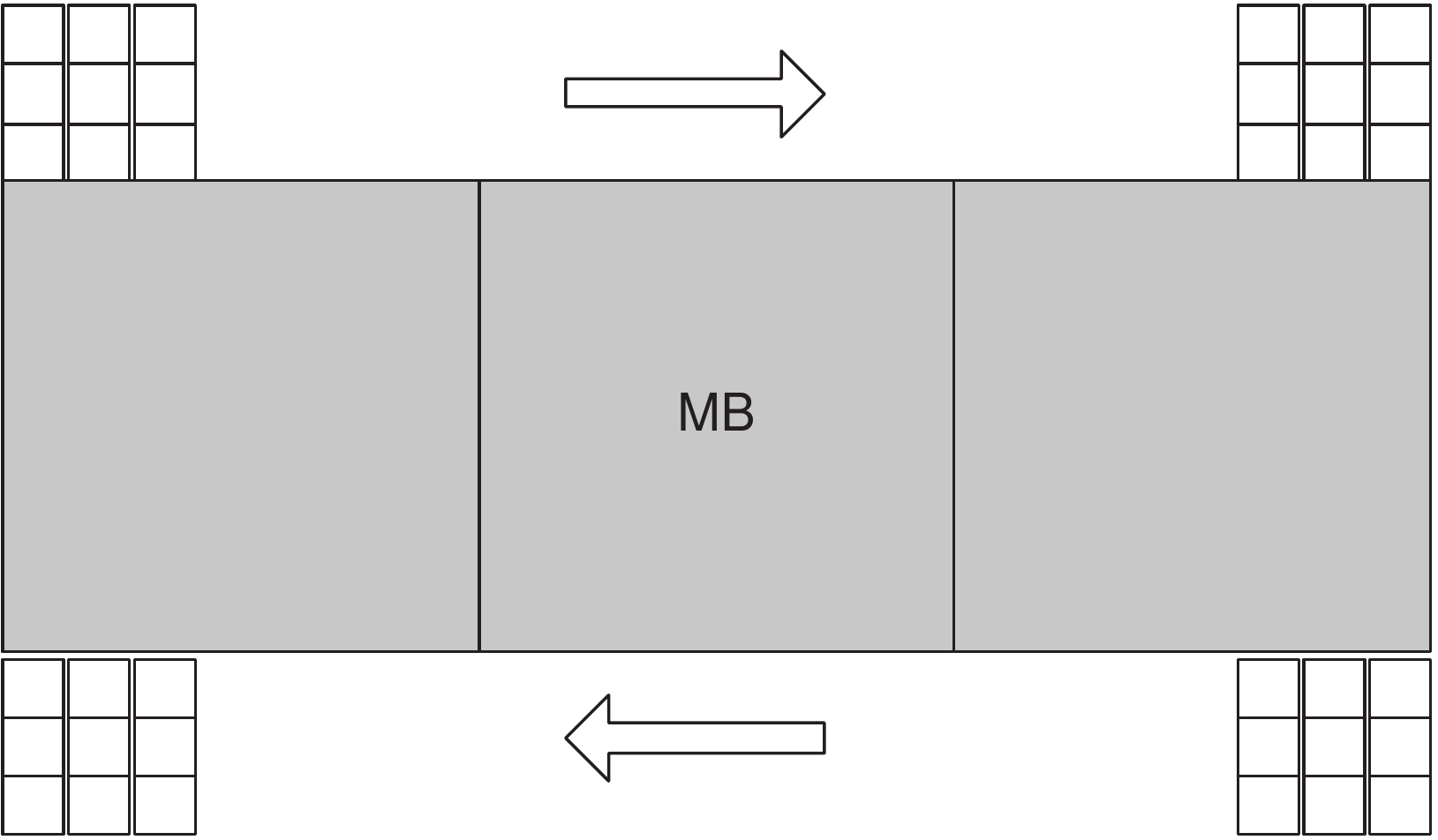}
\label{Fig_Chapter4_ConBlock}
\caption{Left: MB with its surrounding blocks; Middle:Individual and Right: consecutive block loss}
\label{Fig_Chapter4_Blockloss}
\end{figure}

The spatial-domain concealment algorithms edge features play an important role in describing characteristics of content in local image. Therefore, missing image blocks can be restored by extending edges that are present in the surrounding neighborhood so that they pass through the missing block. Fig.\ref{Fig_Chapter4_Blockloss} shows a missing block, MB, with its surrounding neighborhood, NB. In order to represent edge orientations, we introduce a gradient directional vector (GDV), found by the edge structure of a surrounding neighborhood. The pixel regions of NB\lowercase{u} , NB\lowercase{d} , NB\lowercase{l} and NB\lowercase{r} are used in order to obtain redundancy in all the available surrounding blocks. If NB\lowercase{l} and NB\lowercase{r} are not available, the pixel regions of NB\lowercase{ul} , NB\lowercase{dl} , NB\lowercase{ur} and NB\lowercase{dr}  are used instead. The edge features in the error block should be estimated according to the edges in the surrounding available blocks.To perform this classification, a gradient filter is employed for simple and fast detection. In the neighboring blocks shown in Fig.\ref{Fig_Chapter4_Blockloss}, each boundary pixel $ NB(x, y)$ is convolved with the  Sobel masks.The local edge gradient components for the pixel $NB(x, y)$ and  the magnitude and angular direction of the gradient at coordinate (x, y) are computed by

\begin{equation}
\begin{array}{lcl}
S_{x}(x,y)=NB(x,y)\otimes g_{x}(x,y)\\
S_{y}(x,y)=NB(x,y)\otimes g_{y}(x,y)
\label{Equ_Chapter4_SobelConvolution}
\end{array}
\end{equation}
\begin{equation}
\begin{array}{lcl}
M_{s}=\sqrt{S_{x}^{2}(x,y)+S_{y}^{2}(x,y)}\\
P_{\theta }=tan^{-1}\frac{S_{y}(x,y)}{S_{x}(x,y)}
\label{Equ_Chapter4_PhaseMagnitude}
\end{array}
\end{equation}

%%%%%%%%%%%%%%%%%%%%%%%%% Chap4_Fig.23 starts here %%%%%%%%%%%%%%%%%%%%%%%%
%%%%%%%%%%%%%%%%%%%%%%%%% Chap4_Fig.23 ends here %%%%%%%%%%%%%%%%%%%%%%%%%%

%%%%%%%%%%%%%%%%%%%%%%%%% Chap4_Fig.24 starts here %%%%%%%%%%%%%%%%%%%%%%%%
\begin{figure*}
\centering
\includegraphics[width = 4.2in]{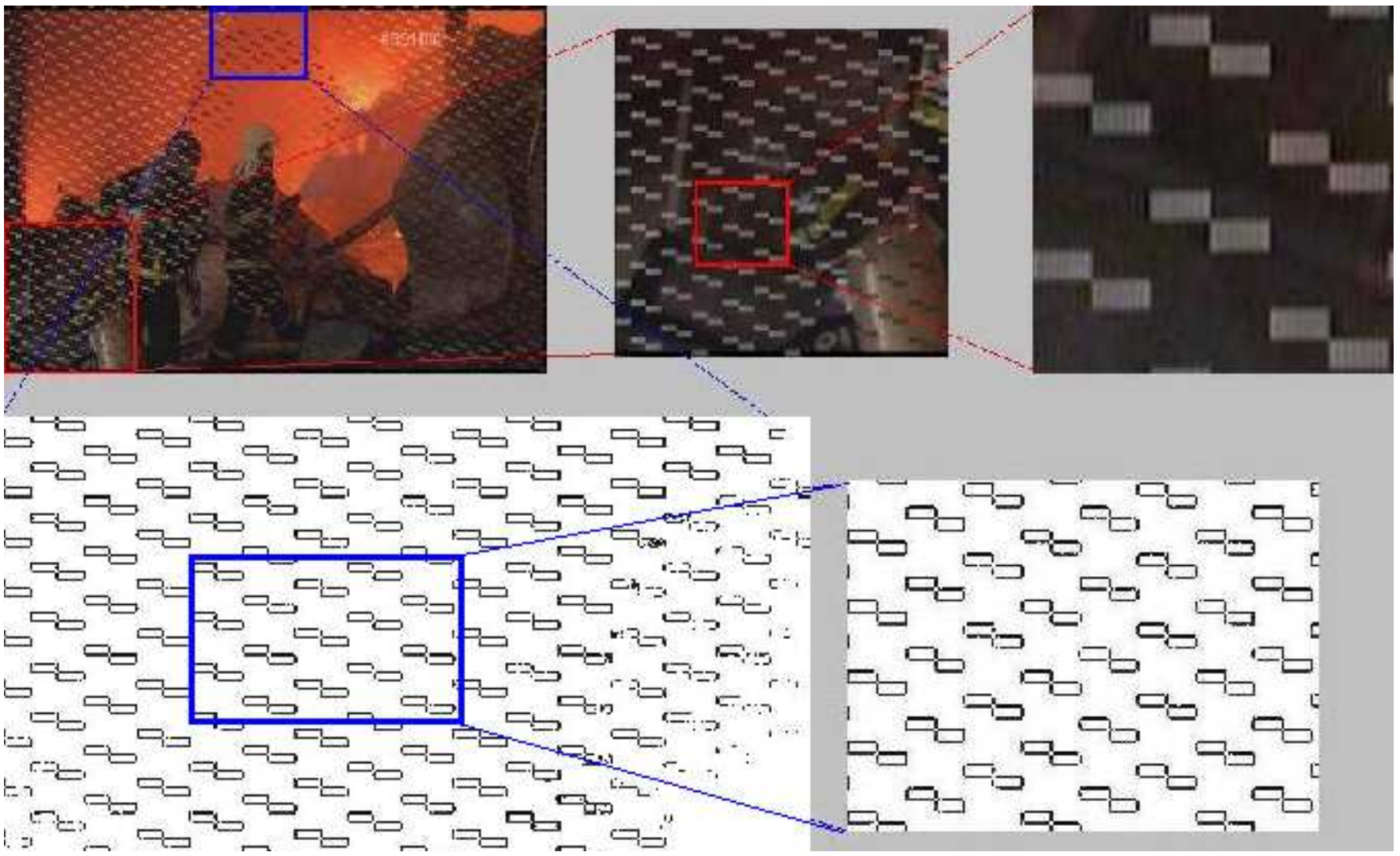}
\label{Fig_Chapter4_Errorblock}
\includegraphics[width = 2.5in]{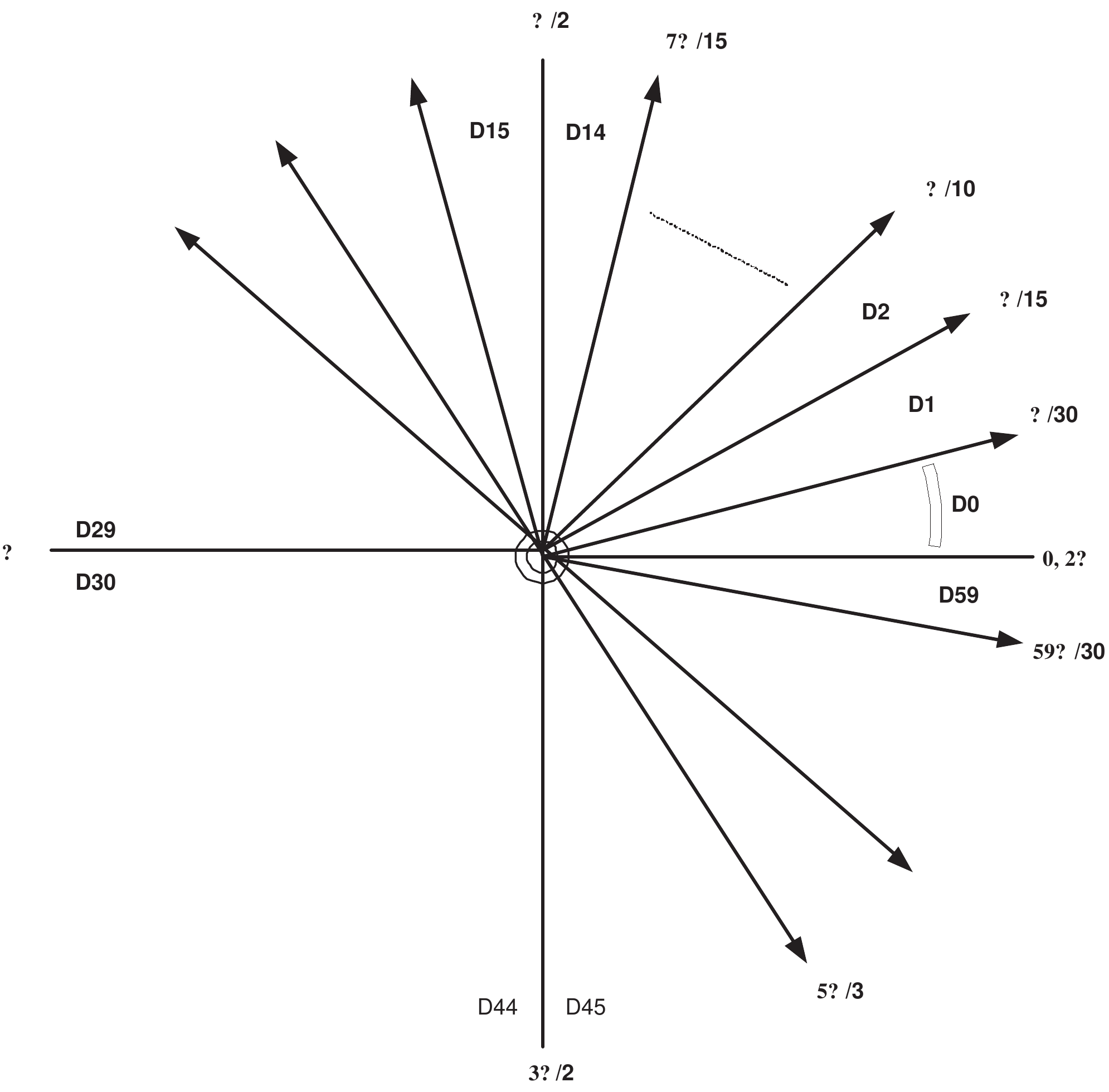}
\label{Fig_Chapter4_Edge60}
\caption{Texture content block of a distorted frame and Edge gradient Classification}
\label{Fig_Chapter4_E}
\end{figure*}
%%%%%%%%%%%%%%%%%%%%%%%%% Chap4_Fig.24 ends here %%%%%%%%%%%%%%%%%%%%%%%%

\par If we analyze the block errors shown in  Fig.\ref{Fig_Chapter4_E}, we found  that they are not like conventional artifacts. The blocks are repeated throughout the whole image and follow a specific pattern. So, we have to accumulate more gradient directions and gain better accuracy. And for that the edge directions are classified into $60$ directions in Fig.\ref{Fig_Chapter4_E}. Each single direction region covers $6^{0}$ and the value of the gradient angle corresponds to one of sixty directional categories equally spaced around $360^{0}$. Also, we define edge magnitude strength (EMS) calculated by equation \ref{Equ_Chapter4_EMS1} for each of the sixty directions. If the edge direction $ P_{\theta }(x,y) $ belong to the $ D_{k} $ direction, the $ EMS_{x}(k) $ and  $ EMS_{y}(k) $ are incremented by an amount of $S_{x}(x,y) $ and $S_{y}(x,y) $, respectively. The gradient of the first quadrant which is $ 0 $ to $ \frac{\pi}{2}$ directions are classified into $D_{0}$ and $D_{14}$ regions respectively.

\begin{equation}
\begin{array}{lcl}
if (\theta (x,y)\in D_{k}) \\
EMS_{x}(D_{k})= EMS_{x}(D_{k})+S_{x}(x,y)\\
EMS_{y}(D_{k})= EMS_{y}(D_{k})+S_{y}(x,y)\\
EMS(D_{k})=\sqrt{EMS^{2}_{x}(D_{k})+EMS^{2}_{y}(D_{k})}
\label{Equ_Chapter4_EMS1}
\end{array}
\end{equation}

In the above step, the edge orientations have been roughly detected. Therefore, a preciseness step for each $ D_{k} $ direction is necessary. We define the gradient direction vector (GDV) to represent a fine and accurate edge orientation as shown in equation \ref{Equ_Chapter4_GDV}.

\begin{equation}
\begin{array}{lcl}
GDV(D_{k})=tan^{-1}\frac{EMS_{y}(D_{k})}{EMS_{x}(D_{k})}
\label{Equ_Chapter4_GDV}
\end{array}
\end{equation}

\subsection{Error Block Classification}

\par In order to conceal each EB with a suitable EC method, the content in the EB should be estimated according to the characteristics in the survived neighboring blocks and also calculation the block orientation and shape is necessary to conceal for the later part. In this work, the content of each EB is estimated  and at the same time it is classified into one of the three categories defined as follows:

\begin{itemize}

\item Uniform block: the gray level of EB may be constant or nearly so. I.e., there is no obvious edge in the block.
\item Edge block: the block locates on the boundary of two or more parts with different gray level. Because the size of block is not large, there are few edges passing through the block and the direction of each edge, in general, is with no or little change.
\item Texture block: both gray level and edge direction varies significantly in the block, so the edge magnitudes of many directions are very strong.

\end{itemize}

\paragraph{Selection of Dominant Directions: By Gradient Magnitude:}

\par The GDVs having a strong $EMS(D_{k})$ can represent real edge directions, but the GDVs having a weak EMS is considered to have insignificant or light directions of uniform blocks. In order to extract the significant vectors from the GDV set, the $EMS(D_{k})$ value in equation \ref{Equ_Chapter4_DomGDV} is used. If the EMS value of a GDV is larger than a certain threshold, $Th_{Ems}$, the corresponding GDV is set as the dominant gradient direction. So, we can classify the spatial error blocks(SEB) depending on their $Th_{Ems}$ response as:

\begin{equation}
\begin{array}{lcl}
EMS_{dom}=MAX_{0}^{k}\left \{ EMS(D_{k}) \right \}\\
Th_{Ems}= EMS_{dom}-Th_{fix}
\label{Equ_Chapter4_DomGDV}
\end{array}
\end{equation}

\begin{equation}
\begin{array}{lcl}
Th_{Ems}>= 0, SEB = EdgeB_{Ems}\\
Th_{Ems} < 0, SEB = UniB_{Ems}
\label{Equ_Chapter4_class1}
\end{array}
\end{equation}

\paragraph{Histogram Accumulation: }

\begin{figure}
\begin{center}
  \includegraphics[scale=0.3]{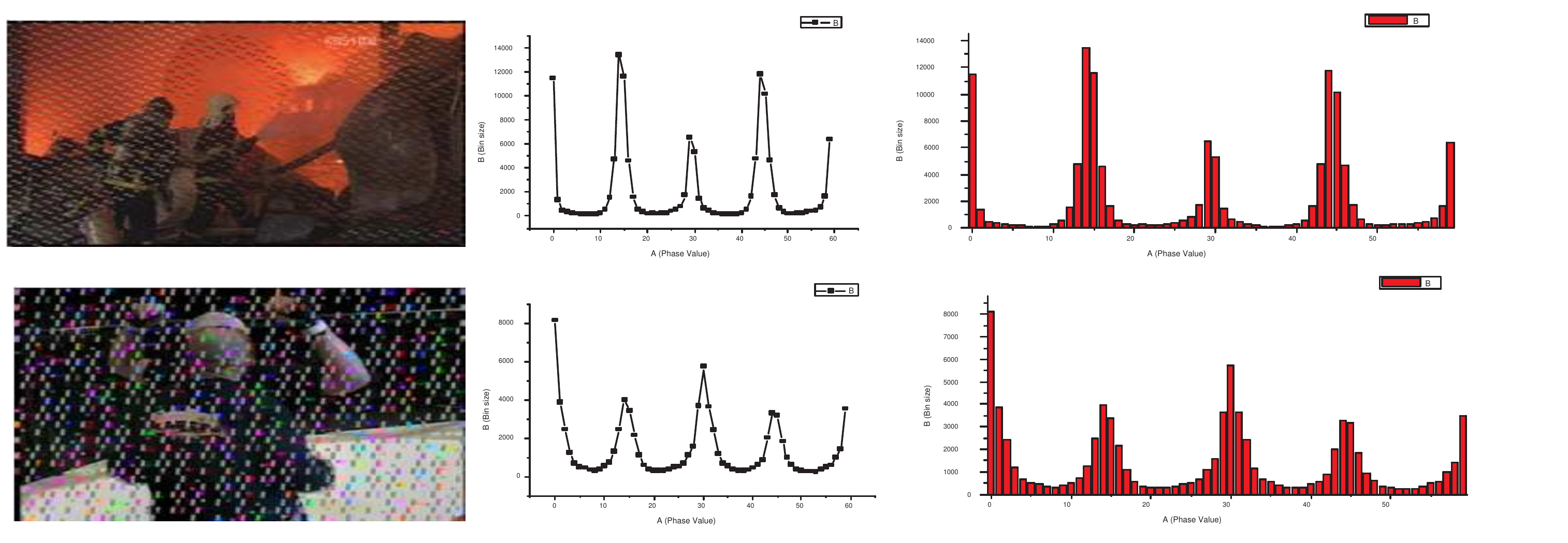}
\end{center}
\caption{Histogram accumulations of images after considering dominant Directions by magnitude}
\label{Fig_Chapter4_histogram}
\end{figure}

The histograms of a digital image with gray levels in the range [0, L-1] is a discrete function $ H_{hist}(r_{k})=n_{k}$, where $r_{k}$ is the $k$th gray level and $n_{k}$ is the number of pixels in the image having gray level $r_{k}$. For our algorithm we use the histogram accumulation of the gradient directions of pixel intensities. Like, if a pixels direction fall in the region of $ D_{14}$ in Fig.\ref{Fig_Chapter4_E}, we assign a phase value of $14$ of that individual pixel. In this way, the gradient of the pixel directions are classified into $D_{0}$ to $D_{59}$ regions with assigning value $0$ to $59$ respectively as shown in Fig.\ref{Fig_Chapter4_histogram}.

%%%%%%%%%%%%%%%%%%%%%%%%% Chap4_Fig.26 starts here %%%%%%%%%%%%%%%%%%%%%%%%
%%%%%%%%%%%%%%%%%%%%%%%%% Chap4_Fig.26 ends here %%%%%%%%%%%%%%%%%%%%%%%%%%

\paragraph{Bin Reduction-By Gradient Phase:}

\par The bin reduction of histogram of gradients is used for classifying the edge blocks and texture blocks. It also can be used for improving the speed and performance of our algorithm. First we analyze which directions dominate in the histogram and then we minimize our bins by considering significant bins that dominate and significantly describe the texture blocks.
\par
\begin{figure}
\centering
\includegraphics[width = 3.5 in ]{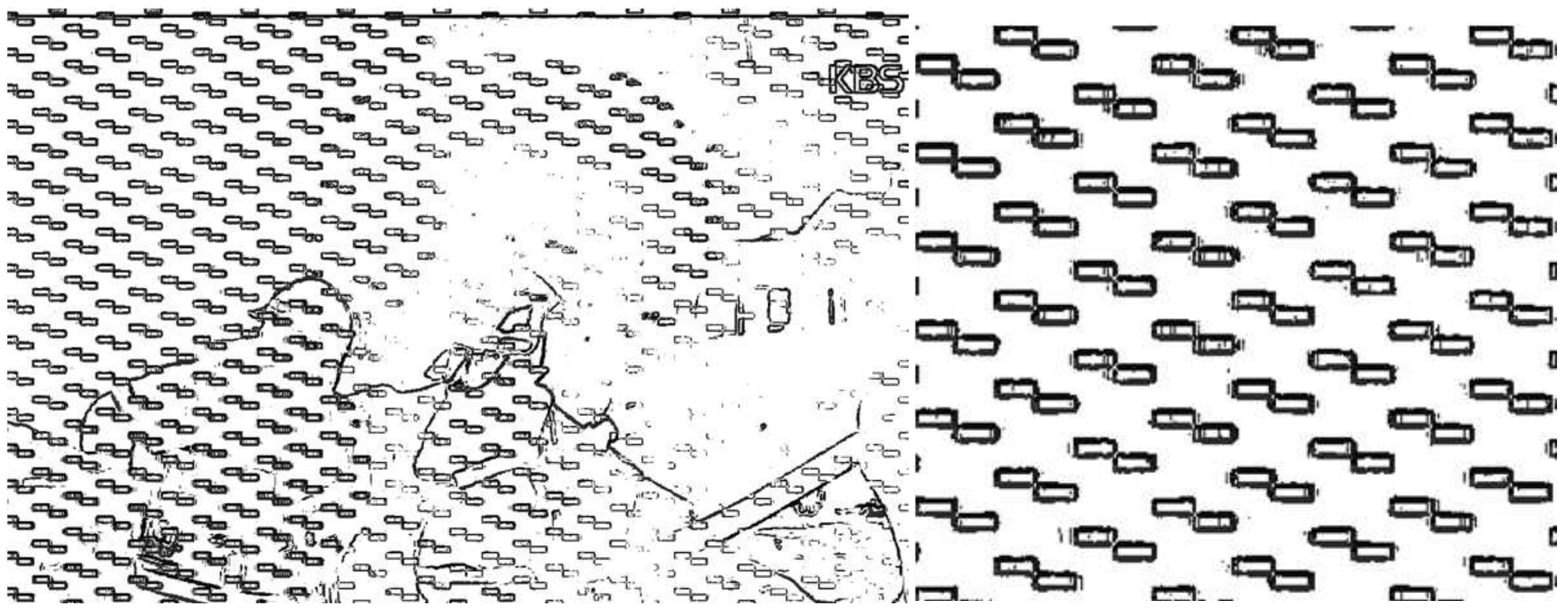}
\label{Fig_Chapter4_bin1}
  \includegraphics[width = 3.5 in]{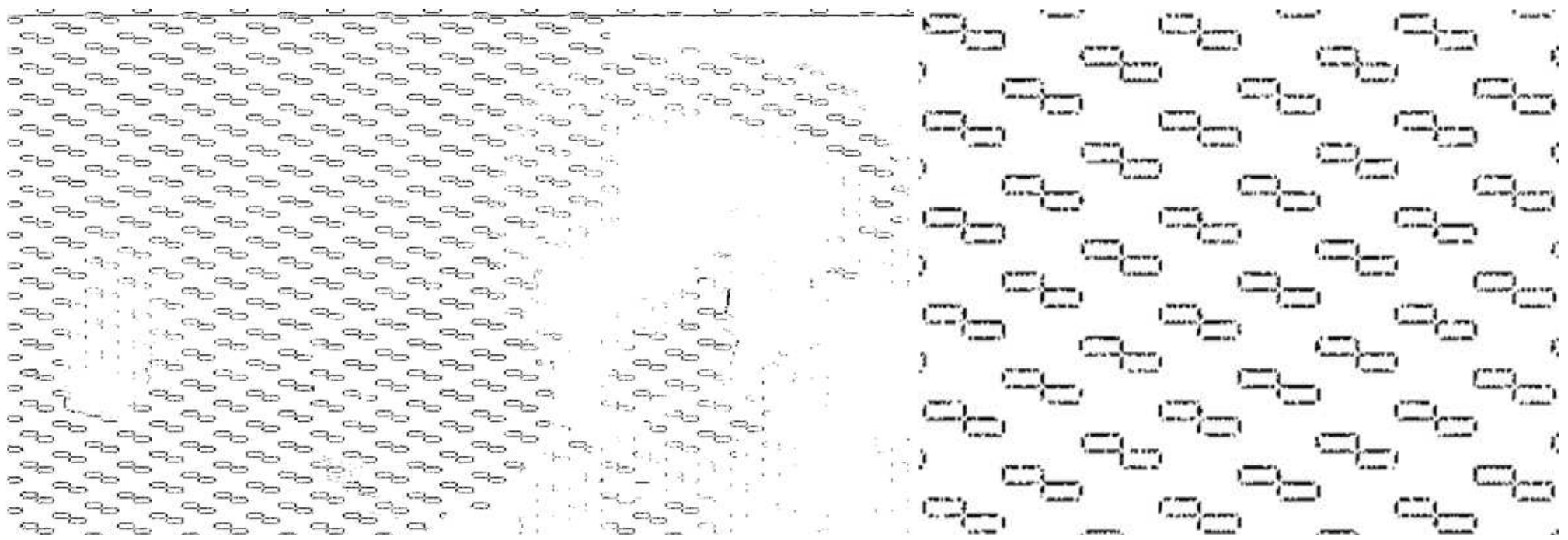}
\label{Fig_Chapter4_bin3}
\caption{Images accumulation considering dominant histogram bins. 60 bins (top) and 12 bins (bottom) }
\label{Fig_Chapter4_binselection}
\end{figure}

 In the Fig.\ref{Fig_Chapter4_binselection}, we have shown the images considering different histogram bins and able to observe that the image includes not only the texture blocks but also the edge blocks. So, to classify the texture blocks from edge blocks we take into account some significant bins. In Fig.\ref{Fig_Chapter4_binselection}, we have shown the images considering $60$ and $12$ significant bins that eliminate the edge blocks form images and keep only the texture blocks.Through observing the dominant histogram bins we reduce the bin accumulation to 12 by considering the bin no: $ 59, 0, 1, 14, 15, 16, 29, 30, 31, 44, 45$ and $46 $. These significant bins not only eliminate the edge portions but also keep track the texture portions and also make our accumulation process faster. As a result we can gain $ 5 $ times faster performance than previous accumulation.

%%%%%%%%%%%%%%%%%%%%%%%%% Chap4_Fig.27 starts here %%%%%%%%%%%%%%%%%%%%%%%%
%%%%%%%%%%%%%%%%%%%%%%%%% Chap4_Fig.27 ends here %%%%%%%%%%%%%%%%%%%%%%%%%%

\paragraph{Bin Reduction Based on Rotation:}

\par If we skim through the different distorted images, we can see that the block patterns are not rotational invariant. For different rotation, the block patterns rotated and their histogram accumulation also changes . In that case, we calculate the phase offset of histogram bins and calculate their rotation . From the offsets we can get the rotation of blocks and use this information in bin reduction process. We will explain the rotation calculation part in section error block analysis.

\subsection{Uniform Block Analysis}
\label{Sec_Chapter4_Uniform}

\par For uniform block analysis , we compare the error block with neighbor blocks that are not corrupted and use neighbor block imitation method to generate an estimated pixel intensity value for the uniform block. The gray levels of pixels in the EB change slowly with the position. So each pixel in the EB can be concealed by linear interpolation using the nearest pixels from the four neighboring blocks along the block boundaries \cite{Suh_con97}. But for our method we use neighbor blocks imitation to estimate block pixels intensity and equation \ref{Equ_Chapter4_ubtb} is applied to imitate block pixel values. Then we took the average values of the pixel intensities of the blocks and get the estimated intensity of every pixels.

\begin{equation}
\begin{array}{lcl}
\sum_{i=1}^{N/2}\sum_{j=1}^{M}UB_{tb}= \sum_{i=1}^{N/2}\sum_{j=1}^{M}B\\
\sum_{i=N/2}^{N}\sum_{j=1}^{M}UB_{tb}= \sum_{i=N/2}^{N}\sum_{j=1}^{M}D\\
\sum_{i=1}^{N}\sum_{j=1}^{M/2}UB_{lr}= \sum_{i=1}^{N}\sum_{j=1}^{M/2}A\\
\sum_{i=1}^{N}\sum_{j=M/2}^{M}UB_{lr}= \sum_{i=1}^{N}\sum_{j=M/2}^{M}C
\label{Equ_Chapter4_ubtb}
\end{array}
\end{equation}

\begin{equation}
\begin{array}{lcl}
\sum_{i=1}^{N}\sum_{j=1}^{M}UB_{Est}= \sum_{i=1}^{N}\sum_{j=1}^{M}\left ( \frac{UB_{tb}+UB_{lr}}{2}\right )
\label{Equ_Chapter4_ubest}
\end{array}
\end{equation}

\subsection{Edge Block Analysis}
\label{Sec_Chapter4_Edge}
\par If the edge of a certain direction is estimated as a strong edge, then the direction is selected as one interpolation direction and a series of one-dimensional interpolation are carried out along the direction to obtain pixel values within the EB. Analysis of the shape and rotation is an important criteria for edge block analysis.

\begin{equation}
\begin{array}{lcl}
DH_{hist}(r_{k})= H_{hist}(r_{k})\otimes M_{hist}(r_{k})
\label{Equ_Chapter4_convolve1}
\end{array}
\end{equation}

\begin{figure}
\centering
  \includegraphics[width=3.0in]{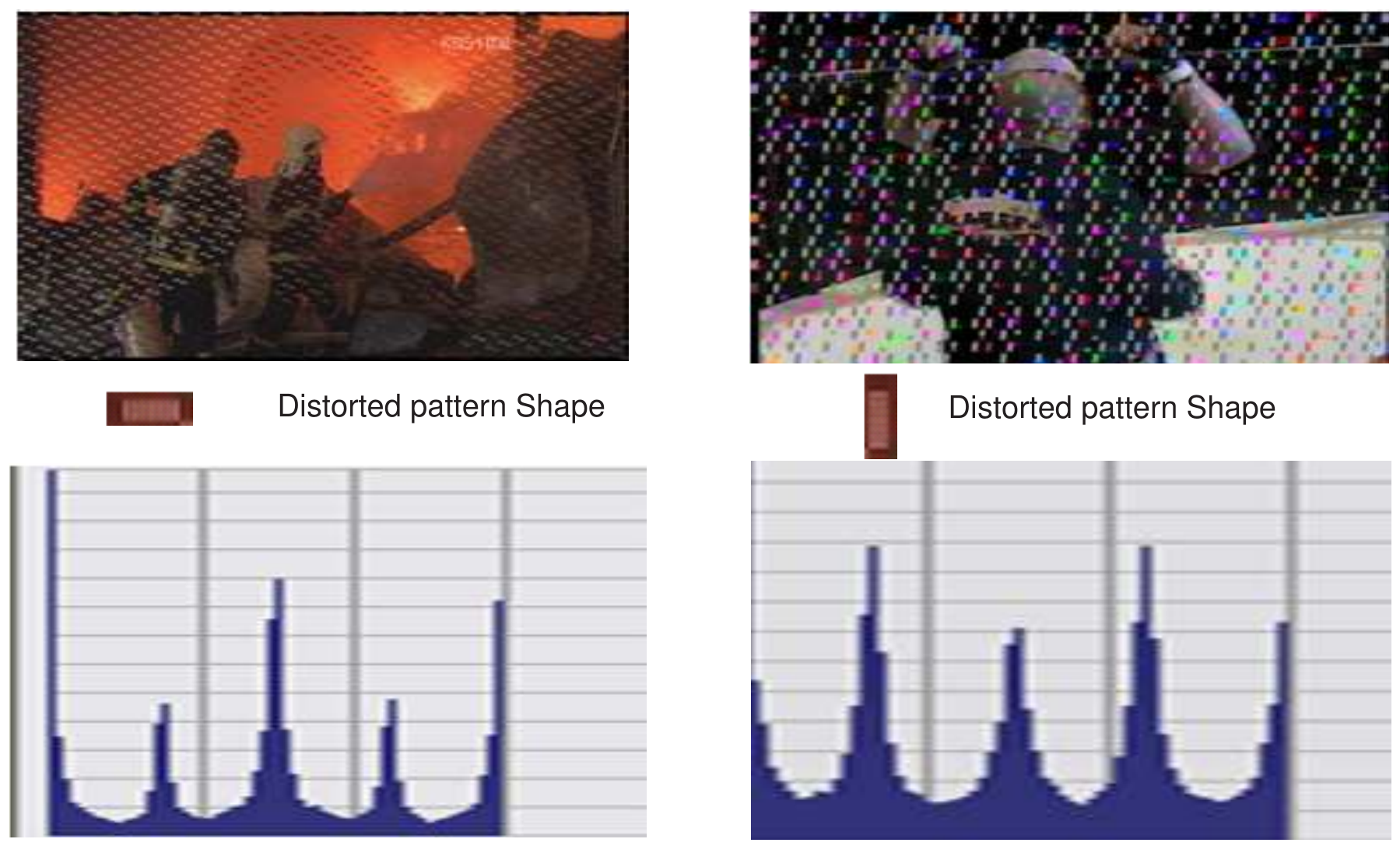}
\label{Fig_Chapter4_edge2}
\includegraphics[width=3.5in]{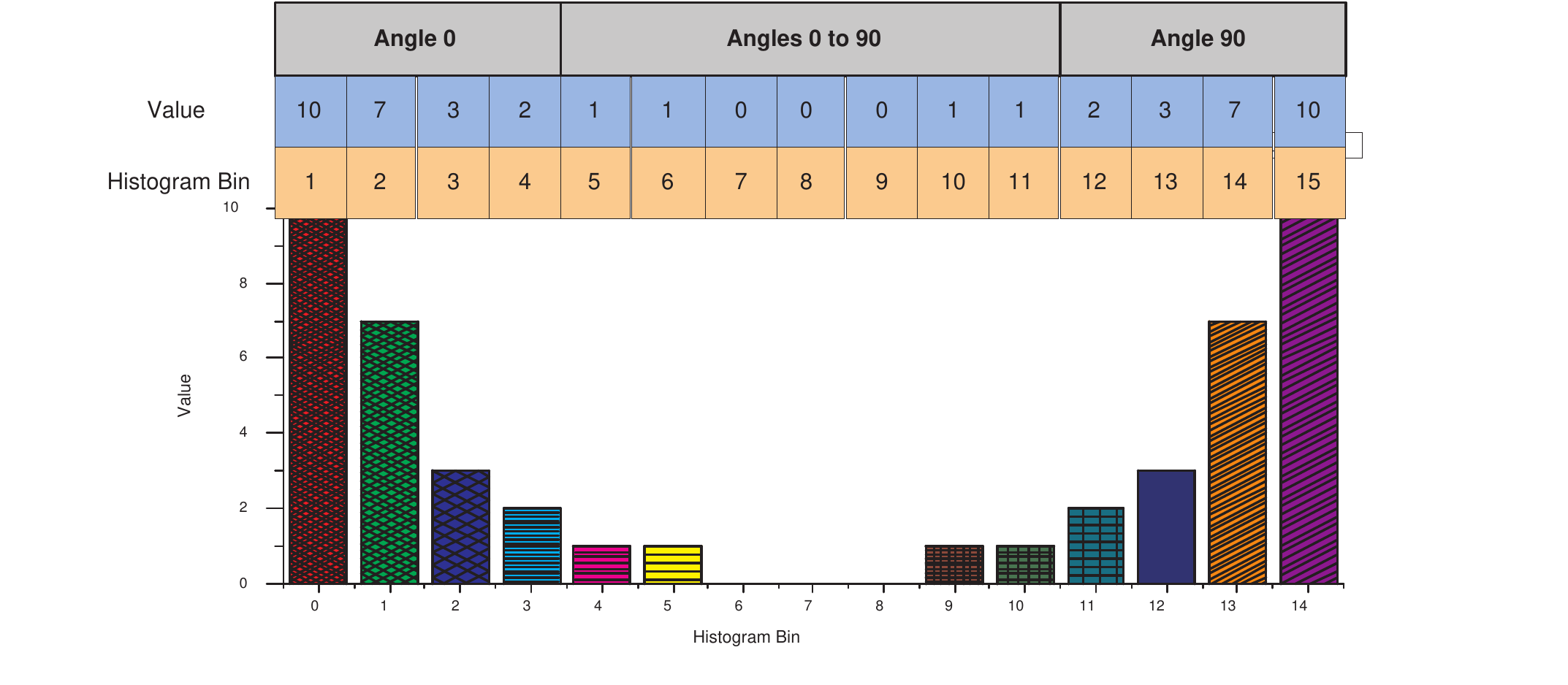}
\label{Fig_Chapter4_mask}
\caption{Histogram, depending on the error block orientation and Histogram mask, convolved with source image histogram}
\label{Fig_Chapter4_edge1}
\end{figure}

In Fig.\ref{Fig_Chapter4_edge1}, we observe that histogram accumulation changes depending on the rotation of the blocks. If the blocks occur horizontally, they emphasize on $0^{0}$ and $180^{0}$ and as a result we get highest accumulation on bin$ 0$ and $30$. On the other hand, if the pattern emphasize vertically, we get highest accumulation on bin $15$ and $45$.With this observation we have come to the decision that, we can also find out the orientation of the artifacts pattern by calculating the phase value from the histogram.

%%%%%%%%%%%%%%%%%%%%%%%%% Chap4_Fig.28 starts here %%%%%%%%%%%%%%%%%%%%%%%%
%%%%%%%%%%%%%%%%%%%%%%%%% Chap4_Fig.28 ends here %%%%%%%%%%%%%%%%%%%%%%%%%%

\begin{equation}
\begin{array}{lcl}
DH_{Phase}(r_{k})= Max\left \{Bin_{0}^{14} \left [ DH_{hist}(r_{k}) \right ] \right \}\\
DH_{offset}(r_{k})=90^{0}-DH_{Phase}(r_{k})
\label{Equ_Chapter4_convolve2}
\end{array}
\end{equation}

\par
\begin{figure}
\begin{center}
  \includegraphics[scale=0.45]{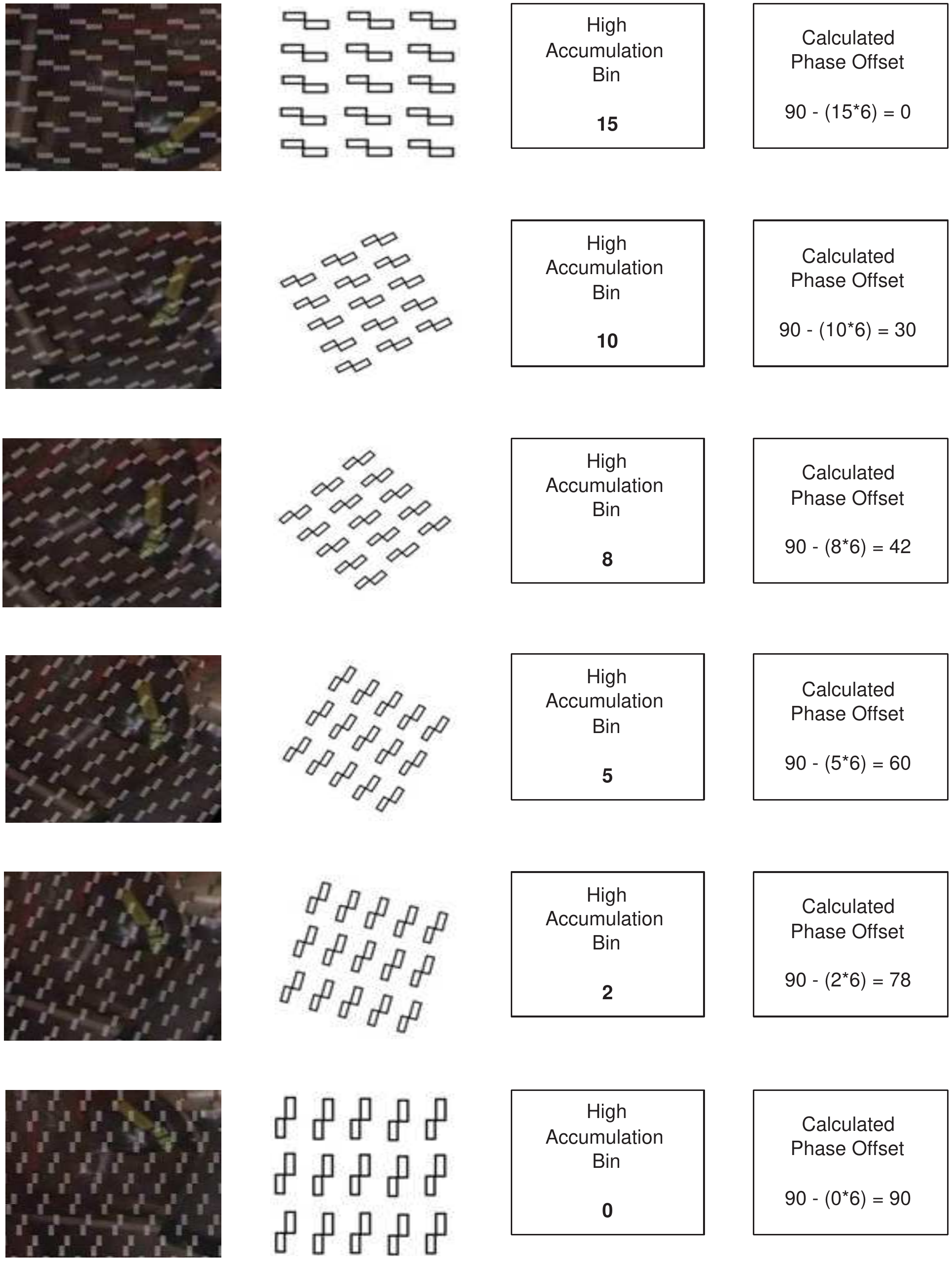}
\end{center}
\caption{Different rotation patterns and their calculated Phase Offset}
\label{Fig_Chapter4_phase}
\end{figure}

 To calculate the phase value or the block orientation from the histogram we will use a circular weighted mask. From the experiment and analysis we have defined the 60 phase circular convolution mask like the Fig.\ref{Fig_Chapter4_edge1}. This 1-D mask is 15 bin long with different values for different bins to give emphasize on significant bins and discard some bins. This mask is  $ M_{hist}(r_{k})$ convoluted with the histogram of the original image histogram  $ H_{hist}(r_{k})$ given in equation \ref{Equ_Chapter4_convolve1} and then shifted on right and convoluted again. In this way we circularly shift the mask and take the accumulation. The high accumulation value will give the phase value with histogram bin. Total $15$ phase shifting needed to find out the orientation of the noise pattern.If the high accumulation is on the bin $5$, then we can say the pattern orientation is , $ 90^{0}- [ 5\times 6 ]= 60^{0}$ .[Because of quantization, the rotation angle will be within 25 to 30 degree], like the Fig.\ref{Fig_Chapter4_phase}. The offset calculated will be used further for concealment.

%%%%%%%%%%%%%%%%%%%%%%%%% Chap4_Fig.30 starts here %%%%%%%%%%%%%%%%%%%%%%%%
%%%%%%%%%%%%%%%%%%%%%%%%% Chap4_Fig.30 ends here %%%%%%%%%%%%%%%%%%%%%%%%%%

\subsection{Texture Content Analysis}
\label{Sec_Chapter4_Discussion}

\begin{figure*}
\begin{center}
  \includegraphics[width=5.5in]{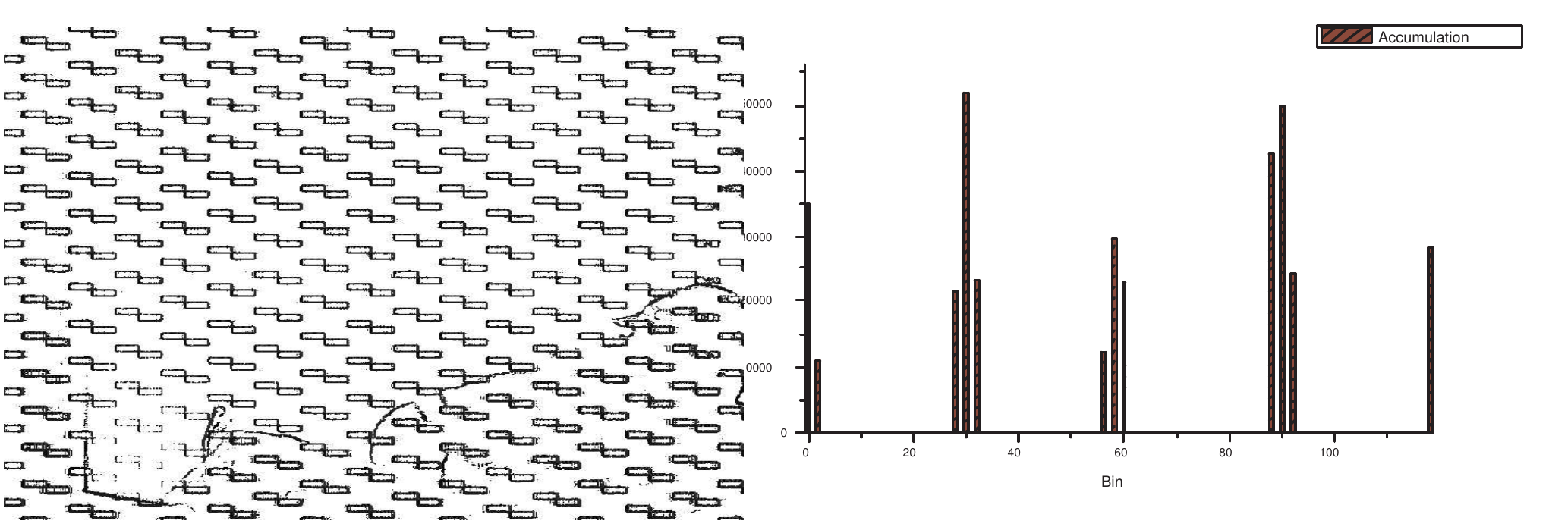}
\end{center}
\caption{Overlapping of the images and their matching score in histogram}
\label{Fig_Chapter4_match}
\end{figure*}

In order to deal with the problem of reconstructing damaged blocks with high details, spatial similarity principles have been applied in existing neighboring blocks and concealment has been performed by searching the best match block \cite{Zeng_conceal99} \cite{Wang_conceal98} \cite{Tsekeri_conceal00}. The best match is the one that minimizes a cost function. To calculate the texture content or to analyze the height and width of the pattern we use matching score. For our purpose we overlap the same gradient phase images and take the histogram accumulation.The total accumulation rises and give emphasize on significant bins shown in Fig.\ref{Fig_Chapter4_match}. The matching score is done by using the equation\ref{Equ_Chapter4_match1}. As we know the total directional vector is between $0 $ to $59$, we can gain highest score of $118$. So, to generate matching score total $118$ histograms are needed. By considering the bin accumulation we can gain the height and width of the pattern. Then by observing significant bins we can get our desired analysis.Through the observations we calculate the height and width of the pattern which are discussed in result section.

%%%%%%%%%%%%%%%%%%%%%%%%% Chap4_Fig.31 starts here %%%%%%%%%%%%%%%%%%%%%%%%
%%%%%%%%%%%%%%%%%%%%%%%%% Chap4_Fig.31 ends here %%%%%%%%%%%%%%%%%%%%%%%%%%

\begin{equation}
\begin{array}{lcl}
D_{match}=D_{i}+ D_{i}
\label{Equ_Chapter4_match1}
\end{array}
\end{equation}

%%%%%%%%%%%%%%%%%%%%%%%%%%%%%%%%%%%%%%%%%%%%%%%%%%%%%%%%%%%%%%%%%%%%%%%%%%%%%%%%%%%%%%%%%%%%%
%%%%%%%%%%%%%%%%%%%%%%%%%%%%%%%%%%%%%%%%%%%%%%%%%%%%%%%%%%%%%%%%%%%%%%%%%%%%%%%%%%%%%%%%%%%%%
%%%%%%%%%%%%%%%%%%%%%%%%%%%%%%%%%%%%%%%%%%%%%%%%%%%%%%%%%%%%%%%%%%%%%%%%%%%%%%%%%%%%%%%%%%%%%
%%%%%%%%%%%%%%%%%%%%%%%%%%%%%%%%%%%%%%%%%%%%%%%%%%%%%%%%%%%%%%%%%%%%%%%%%%%%%%%%%%%%%%%%%%%%%

\section{Experimental Result and Analysis}
\label{Sec_Introduction}

The JPEG image dataset in the LIVE image quality assessment database release 2 \cite{Sheikh_live05} \cite{Sheikh_quality06}\cite{Wang_struct04} and the MPEG-2 video dataset in the LIVE video quality database \cite{Seshadrinathan_object10} are used. The JPEG image dataset includes $29$ color reference images (typically $768 \times¿ 512$ in size) and $204$ JPEG distorted images. The LIVE Video Quality Database uses ten uncompressed high-quality videos with a wide variety of content as reference videos. A set of $150$ distorted videos were created from these reference videos ($15$ distorted videos per reference) using four different distortion types-MPEG-2 compression, H.264 compression, simulated transmission of H.264 compressed bit streams through error-prone IP networks and through error-prone wireless networks. Distortion strengths were adjusted manually taking care of ensuring that the different distorted videos were separated by perceptual levels of distortion. Each video in the LIVE Video Quality Database was assessed by $38$ human subjects in a single stimulus study with hidden reference removal, where the subjects scored the video quality on a continuous quality scale.  Please notice, only the luminance component of each image or video sequence is used for blockiness measurement.

\subsection{ Video Artifact Measure}
\label{Sec_Chapter5_Subjective}
We categorize the experiment section in two parts. First we compare our results with existing JPEG images and then experiment on MPEG-2 video frames.

\subsubsection{Experiments on JPEG Images}
\label{Sec_Chapter5_Subjective}

In the following experiments, we used a number of still images, as well as frames from the test video sequences. These images have different resolutions, ranging from $176\times 144$ to $1920 \times 1080$. We also compared our results with those from other objective quality metrics such as PSNR, the quality metrics $ M_{GBIM}$ of \cite{Wu_impair97}, and the $NR$ quality metrics $S$ of \cite{Wang_perceptual02}. In order to plot all these metrics in the same figure, we scale PSNR by dividing a factor of $5$. According to \cite{Wu_impair97}, there is no defined range for the $M_{GBIM}$ and if $ M_{GBIM}$ values are greater than one, then blocking effect turns out severe. On the other hand, according to\cite{Wang_perceptual02}, the smaller the $S$ is, the greater the severity of the blocking effect is.

%%%%%%%%%%%%%%%%%%% Table 1 Started %%%%%%%%%%%%%%%%%%%%%%%
\begin{table}[!t]
\caption{Pearson Correlation and Spearman for FUB database} \label{Table_Chapter5_FUB}
\begin{center}
\begin{tabular}{|l|c|c|c|}  \hline
 Algorithm 									&	Pearson Correlation 	&	Spearman Correlation \\ \hline
 $Block_{msr}$								&	$-.721$						&	$.685$	\\ \hline
 $ M_{GBIM}$\cite{Wang_perceptual02}&	$-.597$						&	$.584$	 \\ \hline
 $S$\cite{Wu_impair97}					&	$.614	$						&	$.570$	\\ \hline
\end{tabular}
\end{center}
\end{table}
%%%%%%%%%%%%%%%%%%% Table 1 Finished %%%%%%%%%%%%%%%%%%%%%%%
%

%
%%%%%%%%%%%%%%%%%%% Table 2 Started %%%%%%%%%%%%%%%%%%%%%%%
\begin{table}[!t]
\caption{Pearson Correlation and Spearman for LIVE database} \label{Table_Chapter5_Live}
\begin{center}
\begin{tabular}{|l|c|c|c|}  \hline
Algorithm 									&	Pearson Correlation 	&	Spearman Correlation \\ \hline
$Block_{msr}$								&	$-.843$						&	$.838$	\\ \hline
$ M_{GBIM}$\cite{Wang_perceptual02}	&	$-.727$						&	$.925$	 \\ \hline
$S$\cite{Wu_impair97}					&	$.944	$						&	$.937$	\\ \hline
\end{tabular}
\end{center}
\end{table}
%%%%%%%%%%%%%%%%%%% Table 2 Finished %%%%%%%%%%%%%%%%%%%%%%%

\par Table \ref{Table_Chapter5_FUB} shows the Pearson Correlation and Spearman rank order Correction between the proposed blockiness measure and the subjective ratings of QCIF video sequences (obtained from subjective video quality experiments similar to that conducted for the evaluation of the JVT sequences \cite{Vittorio_ISO01}). It can be seen that compared to the metrics of \cite{Wang_perceptual02} and \cite{Wu_impair97}, the $Block_{msr}$ of this paper has a better correlation with subjective test results. Table \ref{Table_Chapter5_Live} shows the Pearson Correlation and Spearman rank-order Correlation between various quality metrics and the subjective ratings of the JEPG database provided by LIVE \cite{Sheikh_live05}. Table \ref{Table_Chapter5_FUB} and Table \ref{Table_Chapter5_Live} show that, our metrics have a comparable correlation with other approaches using subjective data. Additionally, the advantages of our algorithm are that, it is locally adaptive, fast response to blocking artifacts and most of all, it is suitable for real-time implementation. These good technicalities of our algorithm can make it a good choice for practical usage and possibly outweigh the slight drop in correlation values.

\subsubsection{Experiments on MPEG-2 Videos}
\label{Sec_Chapter5_Subjective}
\par The proposed approach can be applied to a video sequence on a frame-by-frame basis. The blockiness measure for a sequence is defined as the mean value of the blockiness measures over all the video frames in the sequence. Testing results on the MPEG-2 video dataset are given in  Table \ref{Table_Chapter5_All} . In the first step we are showing experiment results on our own video datasets. For experimental result, we mention different frames of a video, where different frames have blocking artifacts. We can observe the reliability of our algorithm by comparing $Block_{msr}$ values between them.

\par In Fig.\ref{Fig_Chapter5_fig1} Left column is the color frame and right column is the frame by using kirsch masks. By observing the edge enhanced images we can observe that how the edge direction and magnitude is changed by the blockiness artifacts. Also quantitative measures reflects the artifacts of frames relative to its previous frames but with the same scene. It also shows the another example with different kind of error and the measures shows their significant difference in measuring.Experimental results on the same video dataset using Wu and Yuen's \cite{Wu_impair97}, Vlachos' \cite{Vlachos_blocking00}, Pan et al.'s \cite{Pan_adaptive04}, Perra et al.'s \cite{Perra_blockiness05}, Pan et al.'s \cite{Pan_edge07}, and Muijs and Kirenko's \cite{Mujis_block05} are also reported. From Table \ref{Table_Chapter5_All} we can observe that most of these methods give very satisfactory performance while the proposed outperforms the state of the arts.

%%%%%%%%%%%%%%%%%%% Table 3 Started %%%%%%%%%%%%%%%%%%%%%%%
\begin{table}[!t]
\caption{Test blockiness result using different approaches on the MPEG-2 video dataset} \label{Table_Chapter5_All}
\begin{center}
\begin{tabular}{|l|c|c|c|}  \hline
Approaches 											&	Pearson Corr. 	&	Spearman Corr. &	 RMS Error \\ \hline
Wu and Yuen's				&	$.6344 $					&	$.7365 $					&   $7.1869 $  \\ \hline
Vlachos et al.'s	&	$.5378 $					&	$.7930 $					&   $7.0183 $ \\ \hline
Pan et al.'s 		&	$.6231 $					&	$.6684 $					&   $8.4497 $  \\ \hline
Perra et al.'s&	$.6916 $					&	$.6531 $					&   $8.4357 $  \\ \hline
Pan et al.'s 			&	$.5008 $					&	$.6718 $					&   $8.1979 $  \\ \hline
Muijs and Kirenko's	&	$.7875 $					&	$.6939 $					&   $7.9394 $  \\ \hline
Proposed Method									&	$.8627 $					&	$.7104 $					&   $7.0236 $  \\ \hline
\end{tabular}
\end{center}
\end{table}
%%%%%%%%%%%%%%%%%%% Table 3 Finished %%%%%%%%%%%%%%%%%%%%%%%
%
%%%%%%%%%%%%%%%%%%%%%%%%% Chap5_Fig.32 starts here %%%%%%%%%%%%%%%%%%%%%%%%
\begin{figure}[!t]
\centering
  \includegraphics[width=3.5in]{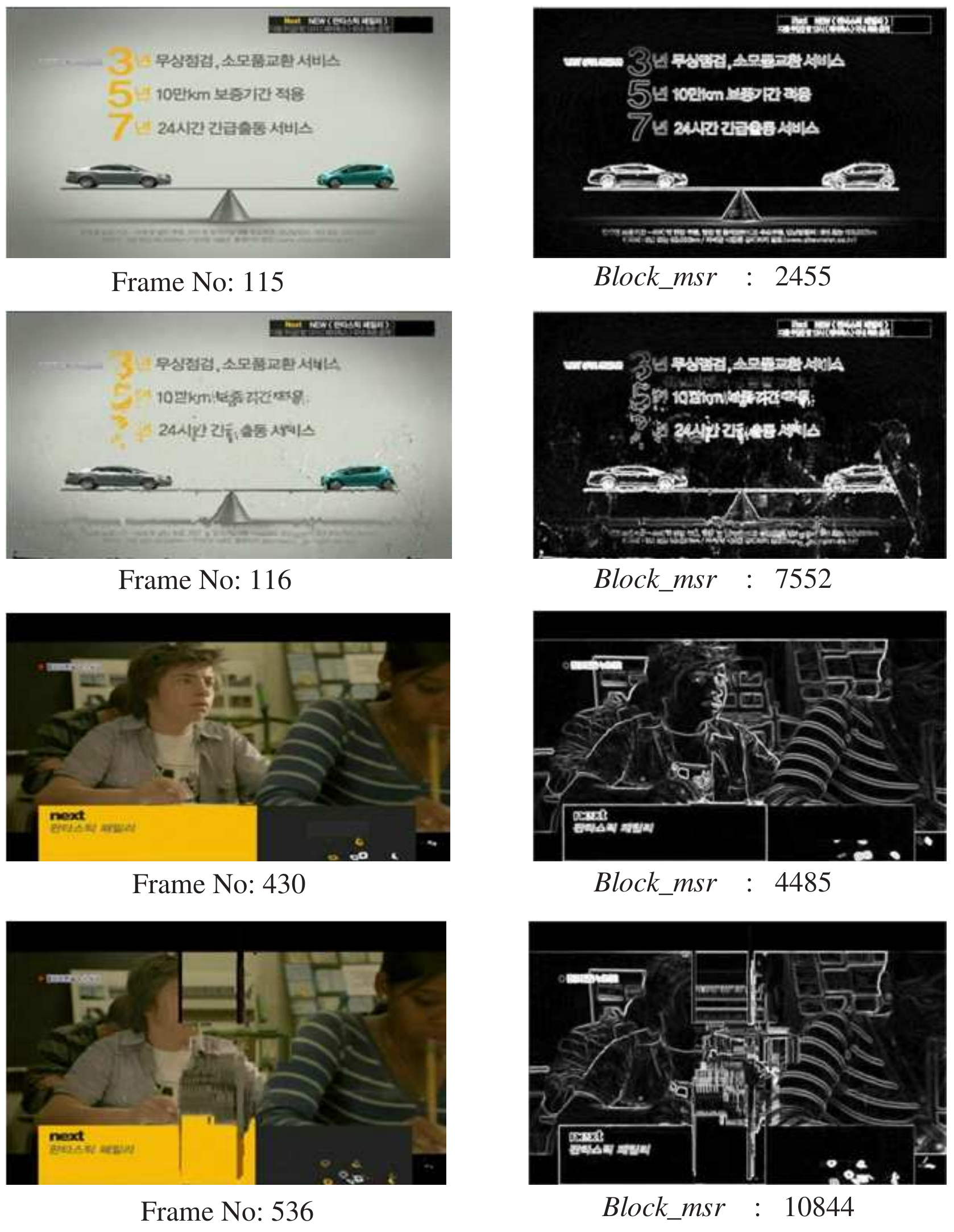}
\label{Fig_Chapter5_fig1}
\includegraphics[width=3.5in]{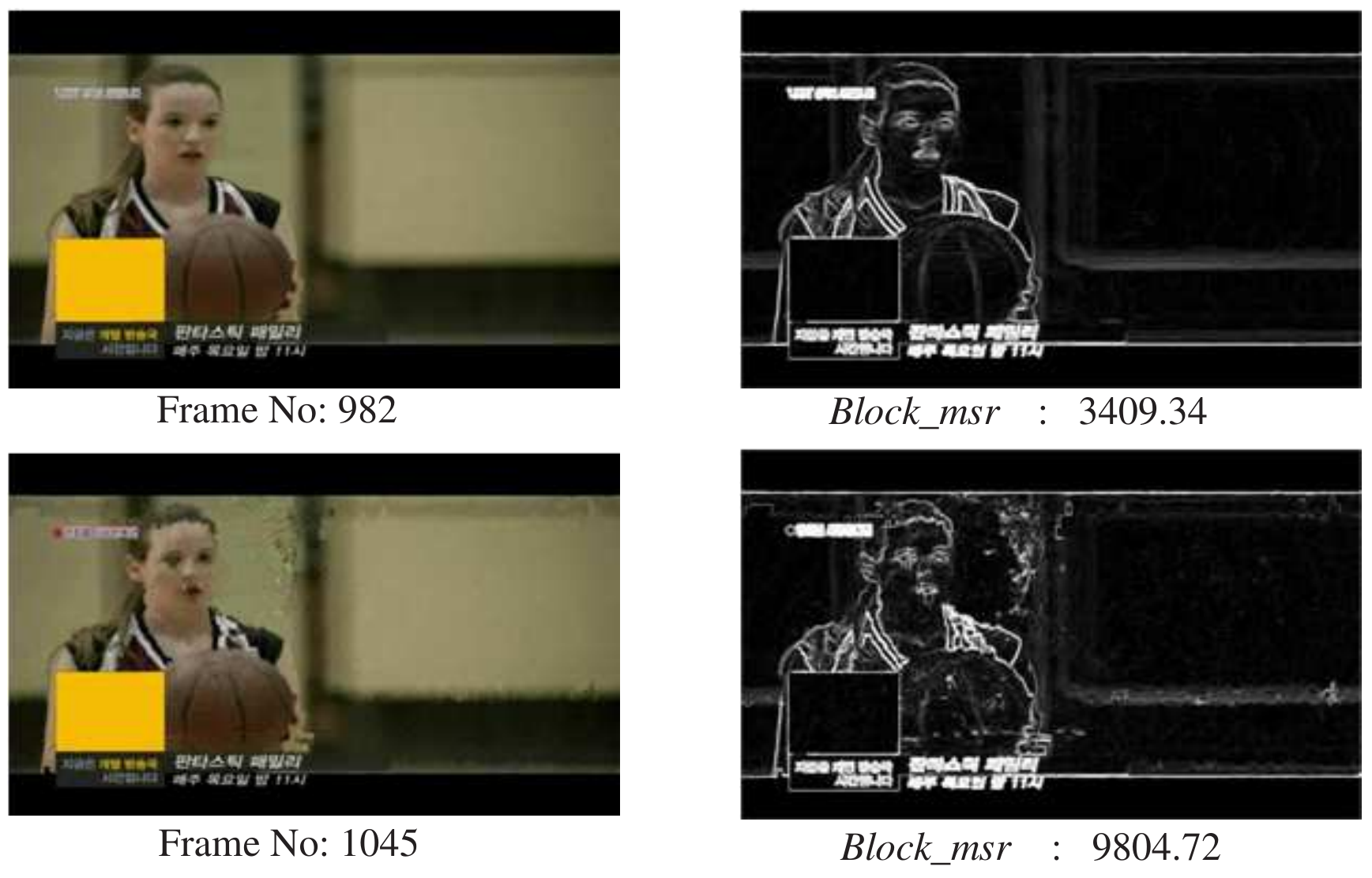}
\label{Fig_Chapter5_fig2}
\caption{Experimental result of different distorted picture quality by using proposed method}
\label{Fig_Chapter5_fig1}
\end{figure}
%%%%%%%%%%%%%%%%%%%%%%%%% Chap5_Fig.32 ends here %%%%%%%%%%%%%%%%%%%%%%%%%%

\subsection{Quantitative Evaluation of Detected Error Frame}
\label{Sec_Chapter5_Quantitative}
\par To compare our algorithm with different approaches the basic measures in accurate detection in general are recall, precision, and efficiency. Recall quantifies what proportion of the correct entities (Number of frames) are detected, while precision quantifies what proportion of the detected entities are correct. Accuracy reflects the temporal correctness of the detected results. Therefore, if we denote by $P$ the distorted frames correctly detected by the algorithm, by $P_{M}$ the number of missed detections (the frames that should have been detected but were not) and by $P_{F}$ the number of false detections (the positions that should not have been detected but were). The equations are given below:

%%%%%%%%%%%%%%%%%%%%%%%%%%%%%%%%%%%%%%%%%%%%%%%%%%%%%%%%%%%%%%%%%%%%%%%%%%
\begin{equation}
\begin{array}{lcl}
Precision = \frac{P }{P+ P_{F}}\\
Recall = \frac{P}{P+ P_{M}}\\
Efficiency = \frac{ Precisioin + Recall }{2}
\label{Equ_Chapter5_EffPrecall}
\end{array}
\end{equation}
%%%%%%%%%%%%%%%%%%%%%%%%%%%%%%%%%%%%%%%%%%%%%%%%%%%%%%%%%%%%%%%%%%%%%%%%%%

%%%%%%%%%%%%%%%%%%%%%%%%% Chap5_Fig.34 starts here %%%%%%%%%%%%%%%%%%%%%%%%
\begin{figure*}
\begin{center}
  \includegraphics[width=5.5in]{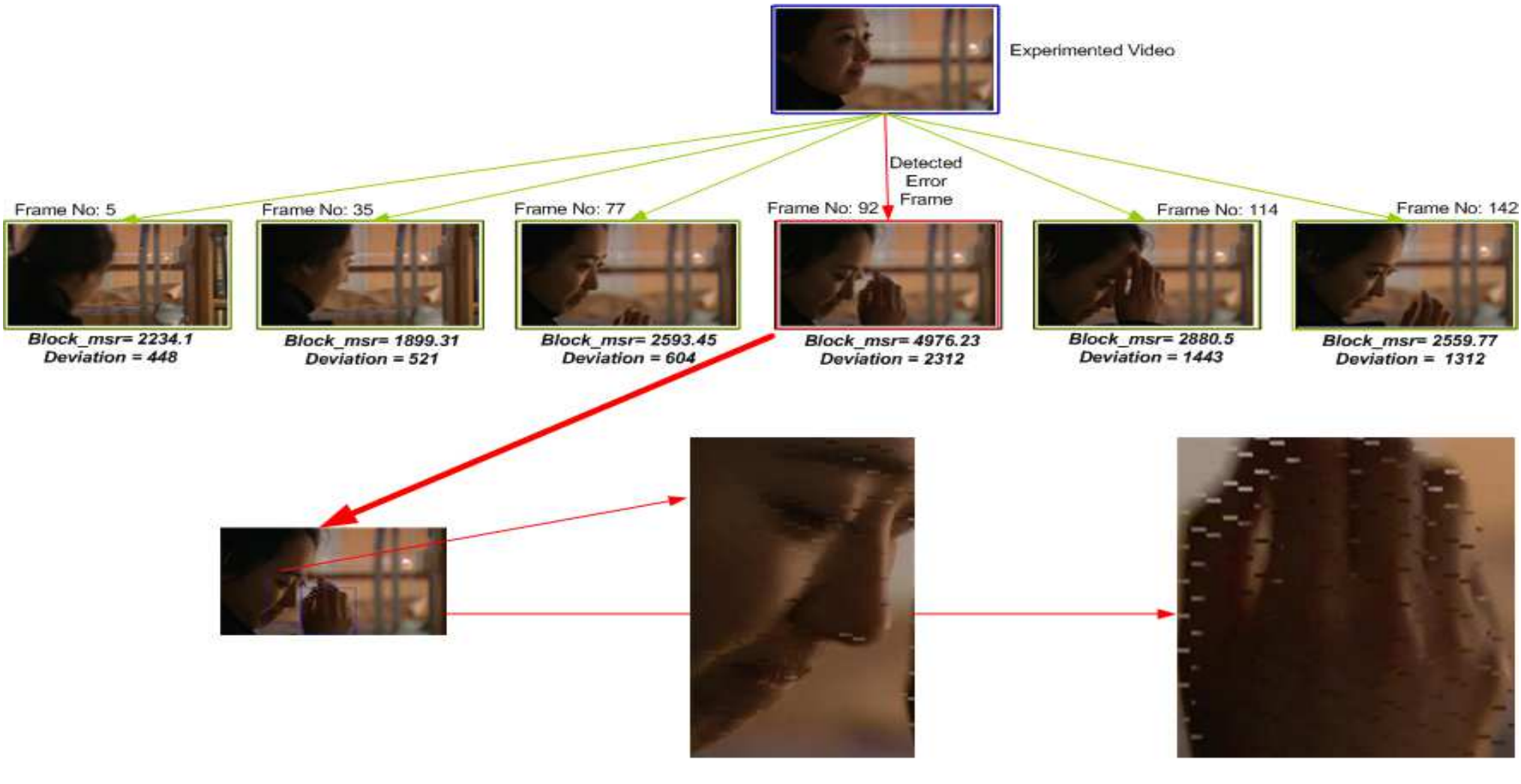}
\end{center}
\caption{ Detection of distorted frames}
\label{Fig_Chapter5_detected}
\end{figure*}
%%%%%%%%%%%%%%%%%%%%%%%%% Chap5_Fig.34 ends here %%%%%%%%%%%%%%%%%%%%%%%%%%
%
\par
\begin{figure}
\begin{center}
  \includegraphics[width=3.0in]{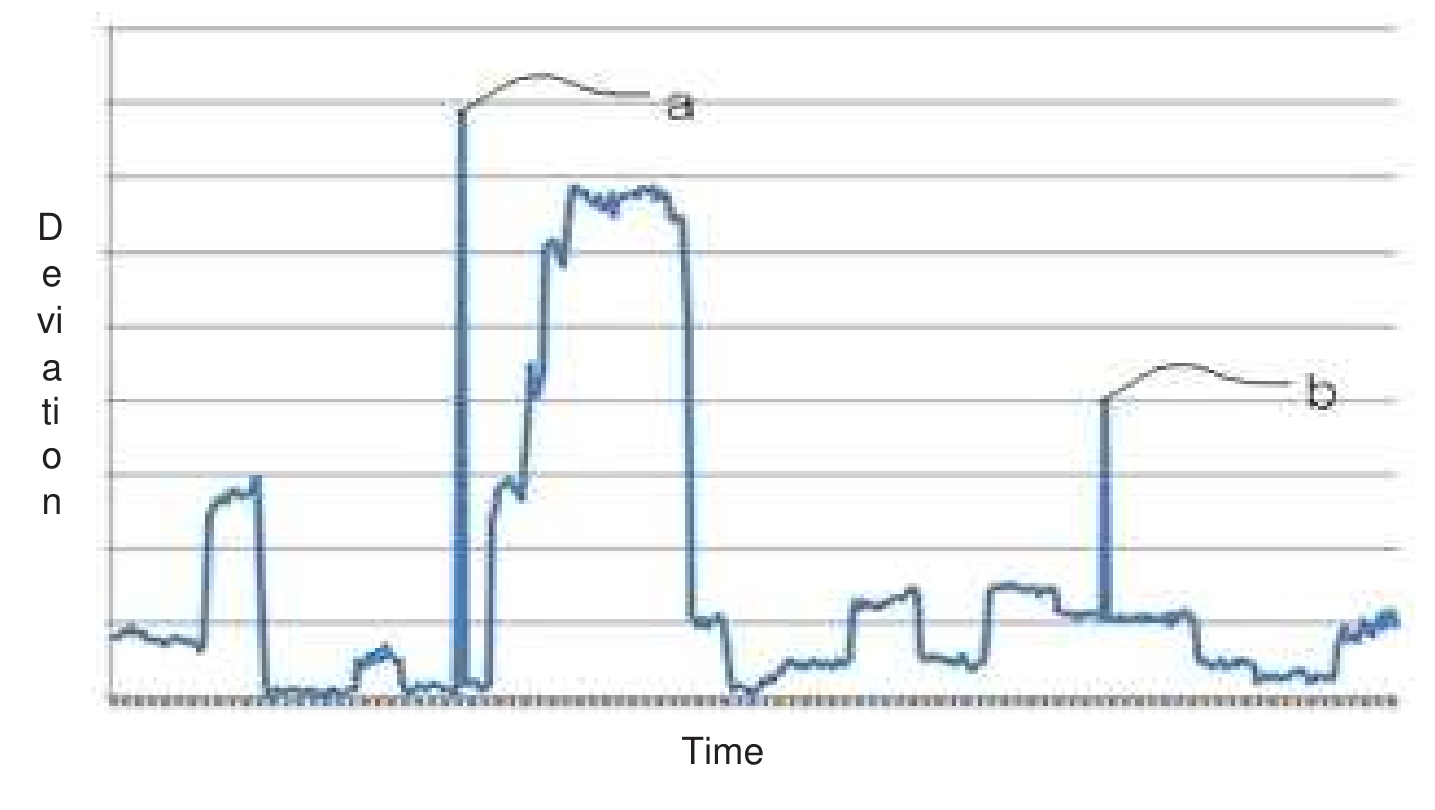}
\end{center}
\caption{ Graph showing the frame deviation graph where a and b are the high edge accumulated distorted frames}
\label{Fig_Chapter5_graph}
\end{figure}

 In our experiment to make the algorithm faster and process the frames in real time, we take into account three previous frames and three next frames to detect distorted frames. So, total seven frames are taken into account for detection process.At first we compute the mean and then calculate the standard deviation of the current frames with previous frames. For our experiment we have used sample video provided by KBS. The video is six seconds video and contain 180 frames. Then after computing the measures we apply equation \ref{Equ_Chapter3_Proposed5}, \ref{Equ_Chapter3_Proposed6} and \ref{Equ_Chapter3_Proposed7}. Then we apply an experimenting criterion to allow the distortions within a certain ratio to detect the defected frame as depicted in equation \ref{Equ_Chapter3_Proposed8}. For test case we use different kind of videos provided by KBS and check it's performance. In Fig.\ref{Fig_Chapter5_detected}, we observe that the criteria function between frame 91, 92 and 93 satisfied our condition. In Fig.\ref{Fig_Chapter5_graph} we have shown a graph that deviation difference between frames with respect to time and a, b are the frames captured as distorted by our algorithm. We also apply this algorithm on LIVE video datasets and gain almost $80$ percent accurate results. This method can detect distorted frames and highly noisy frames from broadcasted videos, accurately.

%%%%%%%%%%%%%%%%%%%%%%%%% Chap5_Fig.35 starts here %%%%%%%%%%%%%%%%%%%%%%%%
%%%%%%%%%%%%%%%%%%%%%%%%% Chap5_Fig.35 ends here %%%%%%%%%%%%%%%%%%%%%%%%%%

%%%%%%%%%%%%%%%%%%% Table 4 Started %%%%%%%%%%%%%%%%%%%%%%%
\begin{table*}
\caption{Comparison of different algorithms showing the detection rate }
\label{Table_Chapter5_detection}
\begin{center}
\renewcommand{\arraystretch}{1.5}
{\footnotesize
\begin{tabular}{|l|cc|cc|cc|cc|}
\hline
 DATASET & \multicolumn{2}{c|}{Wu et al.'s} & \multicolumn{2}{c|}{Pan et al.'s} &
\multicolumn{2}{c|}{Mujis et al.'s} & \multicolumn{2}{c|}{Proposed} \\
LIVE/OCN  & \multicolumn{1}{c}{Recall} & \multicolumn{1}{c|}{Prec.} &
\multicolumn{1}{c}{Recall} & \multicolumn{1}{c|}{Prec.} & \multicolumn{1}{c}{Recall} &
\multicolumn{1}{c|}{Prec.} & \multicolumn{1}{c}{Recall} &
\multicolumn{1}{c|}{Prec.}\\\hline

$Blue Sky $    &     87.01 & 87.02  &   88.31 & 98.27    &  69.83 & 79.85     &   86.35 & 95.40 \\
$Pedestrian $  &     88.88 & 88.03  &   83.34 & 93.31    &  67.29 & 77.24     &   76.74 & 96.52 \\
$River Bed $   &     76.58 & 86.50  &   87.57 & 97.57    &  64.28 & 74.89     &   75.54 & 92.26 \\
$Rush Hour $   &     77.64 & 87.54  &   86.83 & 96.83    &  68.80 & 78.02     &   77.63 & 90.60 \\
$Park Run  $   &     78.08 & 82.05  &   77.35 & 97.32    &  66.20 & 76.23     &   85.47 & 95.49 \\ \hline

$One  $         &     69.44 &  89.28   &     77.77 &  93.33  &     63.89 & 79.31  &   83.33 &  96.77 \\
$Mr. Big  $     &     70.23 &  88.67   &     79.41 &  90.94  &     68.56 & 75.42  &   85.58 &  98.11 \\
$Swim  $        &     66.87 &  85.72   &     84.56 &  96.24  &     65.55 & 78.56  &   88.23 &  95.46 \\
\hline
\end{tabular} }
\end{center}
\end{table*}
%%%%%%%%%%%%%%%%%%% Table 4 Finished %%%%%%%%%%%%%%%%%%%%%%%

\par In the table \ref{Table_Chapter5_detection}, we have shown the detection rate of different algorithms in 100 percent. The quantitative values are given in Recall and precision . Where for test case we have used the LIVE databases given in \cite{Sheikh_live05} LIVE Lab and the OCN databases are provided by Korea Broadcasting System[KBS]. And the comparison shows that our proposed algorithm gives good result when we are dealing with LIVE, which is actually a compression based databases a and outperforms than other when we are dealing with transmission and broadcasting related databases and distortions.

\subsection{Error Block Pattern Analysis }
\label{Sec_Chapter5_Block}

The error block analysis is constructed in two ways. Fist we have shown the  rotation of the blocks is formulated by using rotational convolution mask and then the error pattern analysis by using the overlapping matching scores of the blocks.The idea behind the analysis of rotated pattern is discussed in the previous section.After directional analysis we generate the histogram of the image and then convolve it with our mask given in Fig.\ref{Fig_Chapter4_edge1}. Then we consider the high accumulation bin for our calculation. In the table \ref{Table_Chapter5_pattern}, we have the results of high accumulation bins if the pattern orientation is in different angels. From the table \ref{Table_Chapter5_pattern}, it can be observed that we can accommodate error of $\pm 6^{0}$. Because of the quantization error. As we are quantizing the $360^{0}$ directions in sixty gradient directions and bins, so there will be some error in calculation. To make the system faster, we can compromise the lacking. After calculating the orientation we go the further process of selecting significant bins for deciding the pattern shape and calculation.Also, we have shown the significant bin selection for experiment. The selected bins are then analyzed to calculate the shape of the pattern.

%%%%%%%%%%%%%%%%%%% Table 5 Started %%%%%%%%%%%%%%%%%%%%%%%
\begin{table*}
\caption{Pattern orientation calculation considering histogram bin} \label{Table_Chapter5_pattern}
\begin{center}
\renewcommand{\arraystretch}{1.5}
\begin{tabular}{|l|c|c|c|c|}  \hline
Patternn		&High Bin 	      &	Bin Offset		&	Orientation  & Sig. Bins\\ \hline
$Normal $ 					&$Bin\;15$					&	$15 \times 6 = 90$		&	$90- 90 = 0^{0}$	& $15 \;,\; 30 \;,\; 45\;,\; 0$\\ \hline
$ 30^{0} $ 					&$Bin\; 10$					&	$10 \times 6 = 60$		&	$90- 60 = 30^{0}$	& $10 \;,\; 25 \;,\; 40\;,\; 55$ \\ \hline
$ 45^{0} $ 					&$Bin\; 8$					&	$8 \times 6 = 48$			&	$90- 48 = 42^{0}$	& $8 \;,\; 23 \;,\; 38\;,\; 53$ \\ \hline
$ 60^{0} $ 					&$Bin\; 5$					&	$5 \times 6 = 30$			&	$90- 30 = 60^{0}$	& $5 \;,\; 20 \;,\; 35\;,\; 50$ \\ \hline
$ 75^{0} $ 					&$Bin\; 2$					&	$2 \times 6 = 12$			&	$90- 12 = 78^{0}$	& $2 \;,\; 17 \;,\; 32\;,\; 47$ \\ \hline
$ 90^{0} $ 					&$Bin\; 0$					&	$0 \times 6 = 0$			&	$90- 0 = 90^{0}$	& $0 \;,\; 15 \;,\; 30\;,\; 45$ \\ \hline
\end{tabular}
\end{center}
\end{table*}
%%%%%%%%%%%%%%%%%%% Table 5 Finished %%%%%%%%%%%%%%%%%%%%%%%

%%%%%%%%%%%%%%%%%%%%%%%%% Chap5_Fig.36 starts here %%%%%%%%%%%%%%%%%%%%%%%%
\begin{figure*}
\centering
  \includegraphics[width=3.2in]{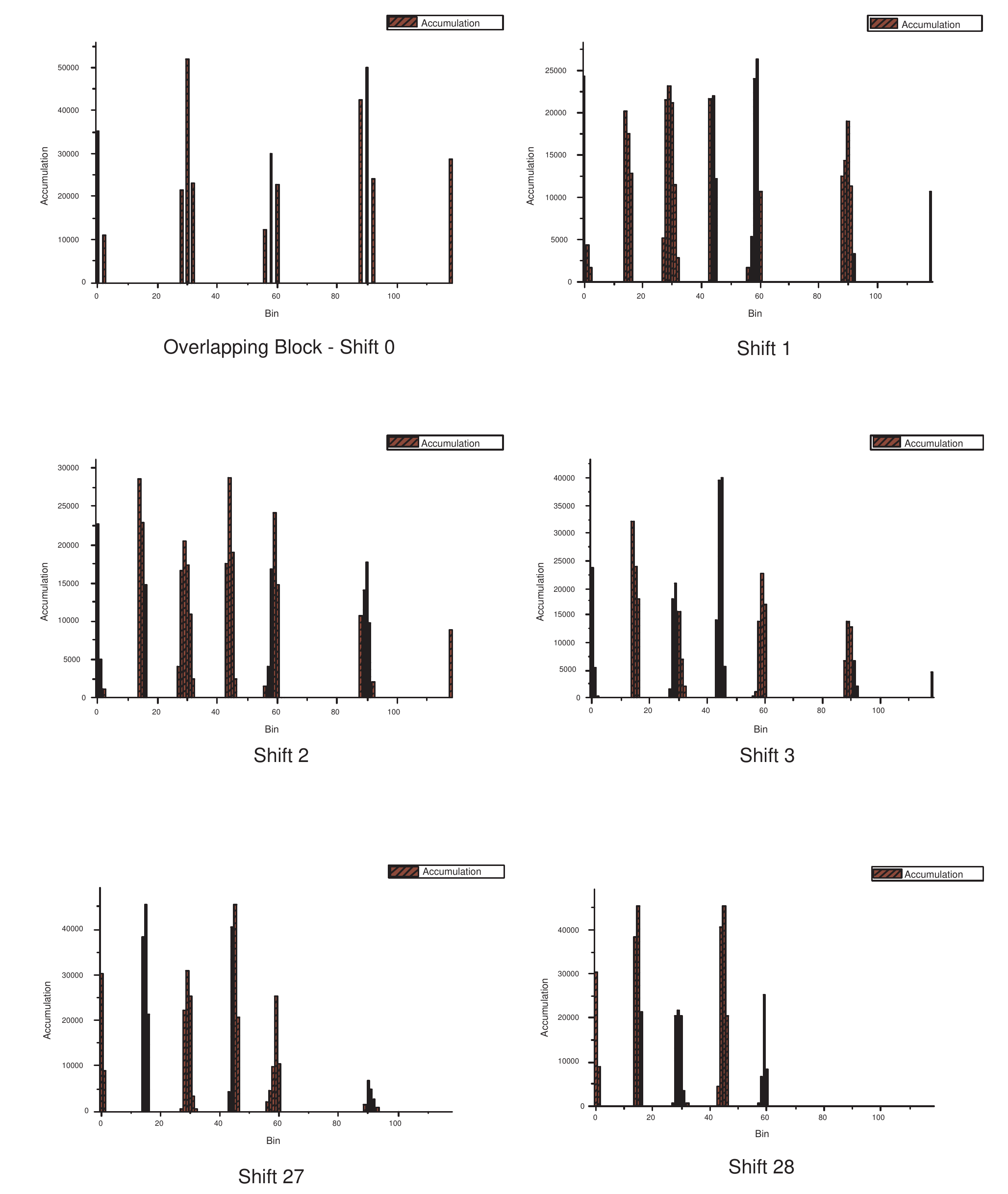}
\label{Fig_Chapter5_shift1}
\includegraphics[width=3.2in]{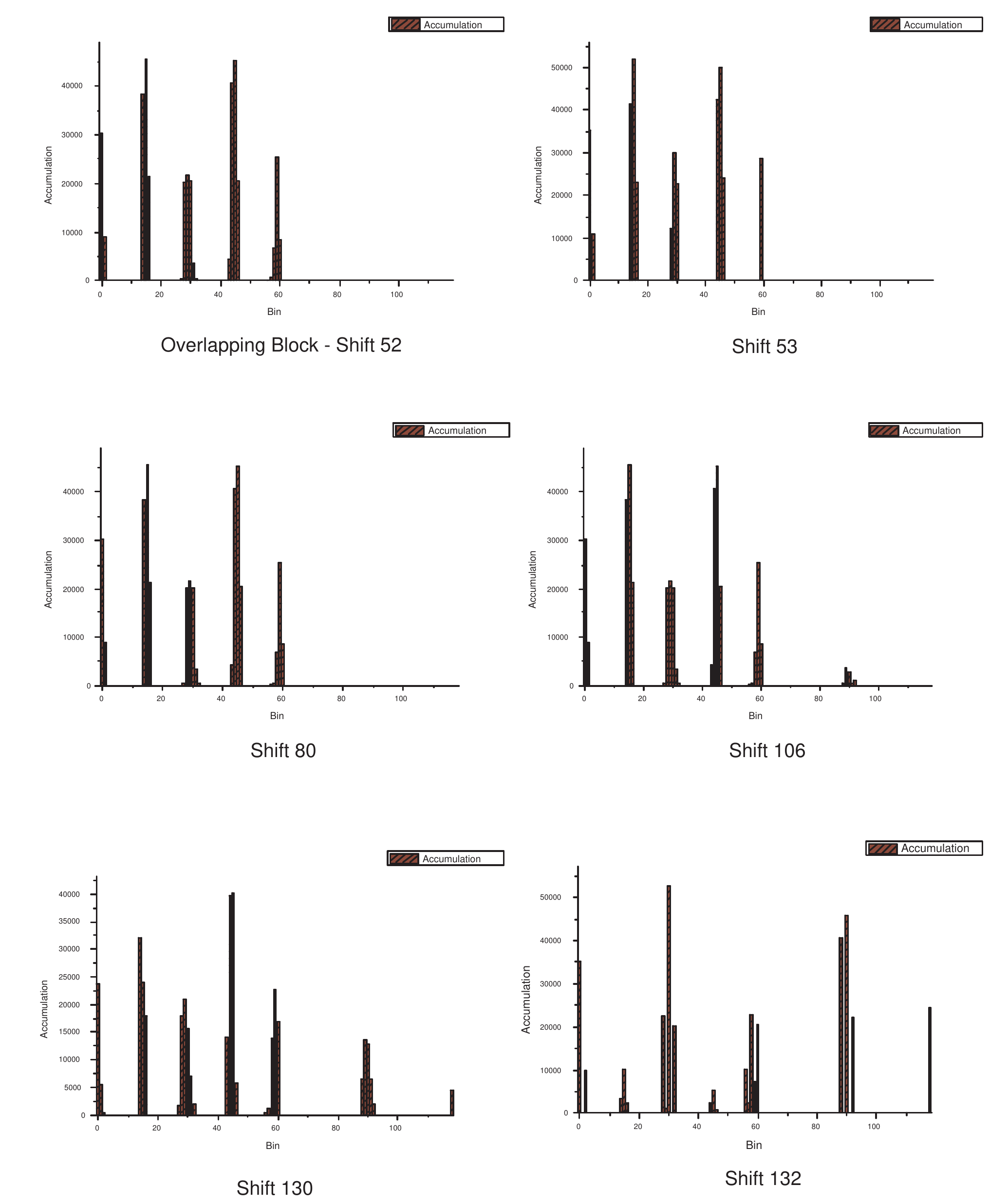}
\label{Fig_Chapter5_shift2}
\caption{ Histogram showing the matching and shifting of overlapping blocks}
\label{Fig_Chapter5_shift}
\end{figure*}
%%%%%%%%%%%%%%%%%%%%%%%%% Chap5_Fig.36 ends here %%%%%%%%%%%%%%%%%%%%%%%%%%

%%%%%%%%%%%%%%%%%%%%%%%%% Chap5_Fig.38 starts here %%%%%%%%%%%%%%%%%%%%%%%%
\begin{figure*}
\begin{center}
  \includegraphics[width=5.5in]{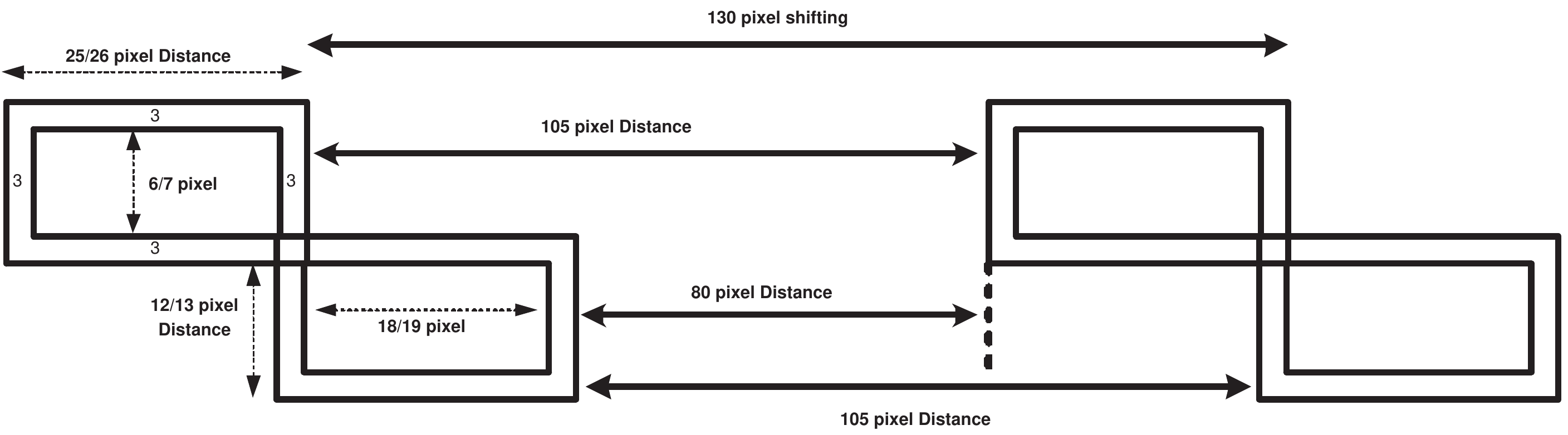}
\end{center}
\caption{ Calculated block pattern}
\label{Fig_Chapter5_pattern}
\end{figure*}
%%%%%%%%%%%%%%%%%%%%%%%%% Chap5_Fig.38 ends here %%%%%%%%%%%%%%%%%%%%%%%%%%

\par As it is tough to get the shape with using global information like histogram, we try to find it out by using pixel/ local information. First we take the same artifacted image. We superimpose one image to another and calculate the matching score. The approach will be Left to right shifting to calculate width and top to bottom to calculate height. At the time of shifting the matching score began to lower and lower. After certain shifting matching score goes to zero. After some shifting the matching score goes up from zero and goes higher. After some shifting the score goes higher (top response when pattern fully superimpose with other pattern) and began to go lower again. We calculate the shifting value to get the height of the pattern. We also can calculate the height of the pattern by using same process. In Figure 25 we have shown different accumulation of matching for different shifting. And show how the graphs satisfy our observations to identify the noise pattern. Here horizontal axis shows the gray scale value of the image and vertical axis shows the phase magnitude accumulation after performing matching. The analysis of our results are shown in Fig.\ref{Fig_Chapter5_shift}.

\par Through the analysis given in the figures we have generated our decisions and observations as following:

\begin{itemize}
\item At Zero shifting the contributing bins are¡¦\\
 $(0, 2, ... ... 28, 30, 32,......, 56,58,60,... 88,90,92,...,128)$
\item After first shifting bin $14, 15, 16,...,43,44,45,46...$ began to dominate and previous bins are shrinking and goes down.
\item After shifting $25/26$ times, Bin $80, 90, 92$ are zero and don't contributing no more.
\item After shifting $53$ times bin 128 also goes down and no more contributing.
\item Then shifting $53$ to $79$ has the same histogram bin accumulation.
\item At shifting $80$ again bin $128$ began to contribute.
\item At shift $105$ bin $88, 90$ and $92$ began to contribute.
\item From shift $80$ to $131$ bins $14, 15, 16$ and bin $43, 44, 45, 46$ began to decrease.
\item At shifting $132$ we gain high accumulation in the contributing bins.
\item At this time the bins contribution is almost same as the initial zero shifting accumulation.
\end{itemize}

High Priority bins to take the decision: 32, (88, 90, 92) and 128  [For Matched Case, High accumulation].Second high priority bins to take decision: (14, 15, 16) and (44, 45, 46) [Accumulation from High (full unmatched) to zero (full matched)].

\section{Conclusion}
\label{conc}

\par The proposed distortion metric is individually calculated as there is a signal discontinuity relative to its local content and its visibility because the masking is locally estimated. Incorporating HVS with this method will make the algorithm more accurate to detect but it will add more computation time to our algorithm which left as future work to gain more visual quality of service. Edge directional information is used instead of using traditional pixel discontinuity along the block boundary. Though if the block has displaced our block does not need to know the exact location of the block to compare with the neighboring blocks. So, our algorithm is invariant to the displacement of block. Even though if the image rotated and in different scale, still we can analyze patterns for concealment. Combining the results in a simple way yields a metric that shows a promising performance with respect to practical reliability, prediction accuracy and computational efficiency.

%%%%%%%%%%%%%%%%%%%%%%%%%%%%%%%%%%%%%%%%%%%%%%%%%%%%%%%%%%%%%%%%%%%%%%%%%%%%%%%%%%%%%%%%%%%%%

\balance
\bibliographystyle{IEEEtran}
\bibliography{ref}
\end{document}